\documentclass[10pt,twocolumn,letterpaper]{article}
\usepackage[table,xcdraw]{xcolor}
\usepackage{float}
\usepackage[pagenumbers]{cvpr} % To force page numbers, e.g. for an arXiv version

\definecolor{cvprblue}{rgb}{0.21,0.49,0.74}
\usepackage[pagebackref,breaklinks,colorlinks,linkcolor=red,citecolor=cvprblue]{hyperref}
\usepackage{amssymb}
\usepackage{mathrsfs}
\usepackage{amsthm,amsmath,amssymb}
\usepackage{tabularx}
\usepackage{cuted}
\usepackage{xcolor}
\usepackage{cancel}
\usepackage{tikz}

\newcommand{\underdashline}[1]{%
  \begin{tikzpicture}[baseline=(a.base)]
    \node[inner sep=0pt] (a) {#1};
    \draw[dashed, line cap=round] 
      ([yshift=-1pt]a.south west) -- ([yshift=-1pt]a.south east);
  \end{tikzpicture}%
}

\def\paperID{2086} % *** Enter the Paper ID here
\def\confName{CVPR}
\def\confYear{2026}

\title{Diversity Has Always Been There in Your Visual Autoregressive Models}

\usepackage{amsmath}
\usepackage{amssymb}
\usepackage{mathtools}
\usepackage{multirow}
\usepackage{makecell}
\usepackage{xcolor}
\usepackage[capitalize]{cleveref}
\crefname{section}{Sec.}{Secs.}
\Crefname{section}{Section}{Sections}
\crefname{table}{Tab.}{Tabs.}
\Crefname{table}{Table}{Tables}
\usepackage{graphicx}
\usepackage{booktabs}
\usepackage{wrapfig}
\usepackage{enumitem}
\usepackage{diagbox}
\usepackage{xspace}

\usepackage{color}
\usepackage{pifont}
\usepackage{multirow}
\usepackage{wrapfig,lipsum}
\usepackage{diagbox}
\usepackage{caption}

\usepackage{diagbox}
\usepackage{amsthm}
\usepackage{multicol}
\usepackage{algorithm}
\usepackage{algorithmic}
\usepackage[ruled,vlined,algo2e]{algorithm2e} 
\usepackage[textsize=tiny]{todonotes}
\usepackage{wrapfig}
\usepackage{color}
\usepackage{colortbl}
\usepackage{anyfontsize}
\usepackage{tikz}
\def\ourmethod{{\textbf{DiverseVAR}}\xspace}

\newcommand{\minisection}[1]{\vspace{0.04in} \noindent {\bf #1}\,}

\usepackage{colortbl}

\newcommand{\lightredrow}{\rowcolor[rgb]{1,0.85,0.86}}

\newcommand{\gray}[1]{{\color{gray}#1}}

\usepackage[capitalize]{cleveref}
\crefname{section}{Sec.}{Secs.}
\Crefname{section}{Section}{Sections}
\crefname{table}{Tab.}{Tabs.}

\newcommand{\circlednum}[1]{%
    \tikz[baseline=(X.base)] {
        \node[draw,circle,inner sep=.3pt] (X) {#1};
    }%
}

\vspace{-10mm} % 紧缩作者与正文之间的间距
\author{
Tong Wang$^{1,2}$ \quad Guanyu Yang$^{1}$ \quad Nian Liu$^{2}$ \quad Kai Wang$^{3}$ \quad
Yaxing Wang$^{4}$ \quad Abdelrahman M. Shaker$^{2}$\\
Salman Khan$^{2}$ \quad Fahad Shahbaz Khan$^{2}$ \quad Senmao Li$^{4,2}$\\
$^{1}$Southeast University \quad
$^{2}$MBZUAI \quad
$^{3}$City University of Hong Kong \quad
$^{4}$Nankai University 
}
\vspace{-10mm} % 紧缩作者与正文之间的间距

\begin{document}
\twocolumn[{
\renewcommand\twocolumn[1][]{#1}
\maketitle
\begin{center}
\includegraphics[width=\linewidth]{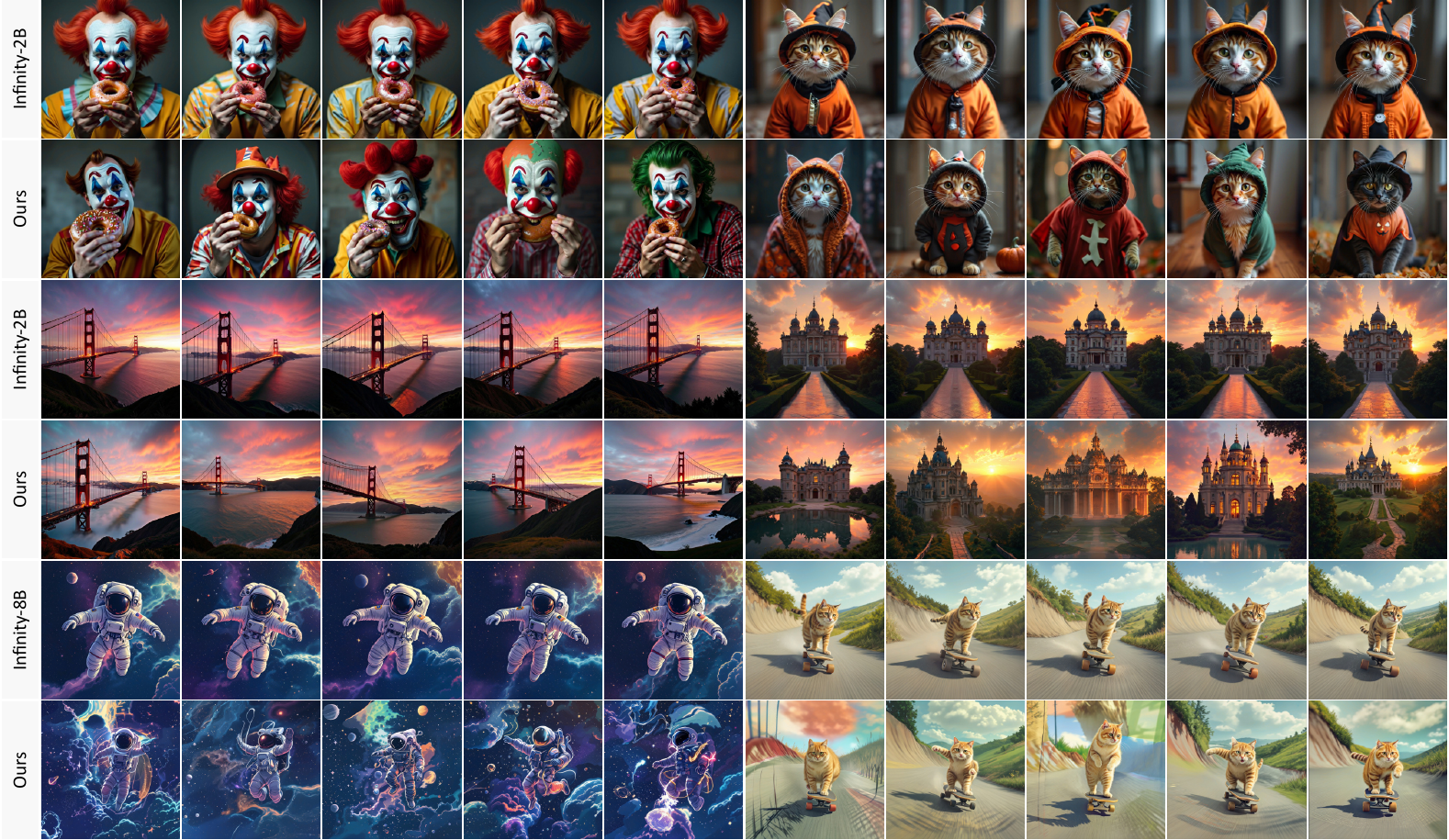}
   \vspace{-1.5\baselineskip}
    \captionof{figure}{Multiple generation samples from the vanilla VAR models (\textit{1st} and \textit{3rd} rows) and our \textit{DiverseVAR} (\textit{2nd} and \textit{4th} rows). While vanilla VAR models suffer from the diversity collapse, our method generates more diverse outputs while maintaining image–text alignment. The text prompts used are as follows: ``A man in a clown mask eating a donut'', ``A cat wearing a Halloween costume'', ``Golden Gate Bridge at sunset, glowing sky, ...'', ``A palace under the sunset'', ``A cool astronaut floating in space'', and ``A cat riding a skateboard down a hill''.
    }
    \label{fig:teaser}
\vspace{0.6\baselineskip}

%% prompts:
% A man in a clown mask eating a donut.
% A cat wearing a halloween costume.
% Golden Gate Bridge at sunset, glowing sky, dramatic perspective, vivid colors.
% A palace under the sunset.
% A cool astronaut floating in space.
% A cat riding a skateboard down a hill.
\end{center}
}]
\maketitle

\begin{abstract}
%Visual Autoregressive (VAR) modeling has attracted significant attention for its novel next-scale prediction methods, offering distinct advantages in inference efficiency and image quality over traditional multi-step autoregressive (AR) and diffusion models.
% However, despite the improved inference efficiency, VAR models face the issue of diversity collapse in the generated outputs, similar to how diffusion models experience this issue when using their few-step distilled versions.
% In this paper, we introduce \textbf{DiverseVAR}, a simple yet effective approach designed to unleash the generative diversity of VAR models without any additional training. 
% In particular, our analysis reveals that the pivotal component of the feature map plays a critical role in governing diversity formation at early scales.
% By suppressing the pivotal component of the model input and augmenting the pivotal component of the model output, \textbf{DiverseVAR} effectively unleashes the inherent generative diversity of VAR models while maintaining high fidelity in the generated results. Furthermore, our results indicate that \textbf{DiverseVAR} effectively unleashes the generative diversity of the model, while incurring only a minimal 0.03 reduction on GenEval.
Visual Autoregressive (VAR) models have recently garnered significant attention for their innovative next-scale prediction paradigm, offering notable advantages in both inference efficiency and image quality compared to traditional multi-step autoregressive (AR) and diffusion models. 
However, despite their efficiency, VAR models often suffer from the diversity collapse i.e., a reduction in output variability, analogous to that observed in few-step distilled diffusion models.
In this paper, we introduce \textbf{DiverseVAR}, a simple yet effective approach that restores the generative diversity of VAR models without requiring any additional training.
Our analysis reveals the pivotal component of the feature map as a key factor governing diversity formation at early scales.
By suppressing the pivotal component in the model input and
% augmenting 
amplifying it in the model output, \textbf{DiverseVAR} effectively unlocks the inherent generative potential of VAR models while preserving high-fidelity synthesis.
Empirical results demonstrate that our approach substantially enhances generative diversity with only neglectable performance influences.
% Our code will be publicly released at 
\href{https://github.com/wangtong627/DiverseVAR}{https://github.com/wangtong627/DiverseVAR}

% The ABSTRACT is to be in fully justified italicized text, at the top of the left-hand column, below the author and affiliation information.
% Use the word ``Abstract'' as the title, in 12-point Times, boldface type, centered relative to the column, initially capitalized.
% The abstract is to be in 10-point, single-spaced type.
% Leave two blank lines after the Abstract, then begin the main text.
% Look at previous \confName abstracts to get a feel for style and length.
\end{abstract}    
\section{Introduction}
\label{sec:intro}
% Diffusion models~\cite{rombach2022high, NEURIPS2020_4c5bcfec, song2021scorebased} have demonstrated remarkable capabilities in image generation through their denoising paradigm, albeit at the cost of requiring dozens to hundreds of inference steps. Distillation methods have been used to distill multi-step diffusion models into few-step versions~\cite{luo2023latent, sauer2023adversarial, lin2024sdxllightning, dao2024swiftbrush}, enabling faster inference. However, this often suffers from diversity collapse under multiple samplings~\cite{kang2024distilling, li2025one}.
%
% Recently, autoregressive (AR) models~\cite{lee2022autoregressive,razavi2019generating,sun2024llamagen,yu2022scaling} have emerged as a powerful alternative to diffusion models, demonstrating competitive performance in visual generation~\cite{sun2024llamagen,wang2025simplear}. In contrast to diffusion models, AR models follow a next-token prediction paradigm that involves a large number of sequential decoding steps.
Autoregressive (AR) models~\cite{lee2022autoregressive,razavi2019generating,sun2024llamagen,yu2022scaling} have emerged as a powerful next-token prediction paradigm~\cite{sun2024llamagen,wang2025simplear}, achieving competitive results in visual generation but suffering from substantial generation latency caused by their numerous sequential decoding steps.
% despite involving numerous sequential decoding steps, resulting in substantial generation latency.
% However, despite their ability to generate high-quality images, both diffusion and AR models suffer from time-consuming and computationally expensive inference due to their multi-step sampling processes.
To alleviate the latency, pioneering studies investigated various representation patterns~\cite{huang2025nfig,pang2024patchpredictionautoregressivevisual,jang2024lantern} and parallel decoding~\cite{chang2022maskgit,teng2024accelerating}.
Recently, unlike conventional next-token prediction in AR models, visual autoregressive (VAR) models adopt a next-scale prediction paradigm~\citep{tian2024visual,Infinity}, achieving efficient and high-quality image generation within approximately ten scale steps.
% Despite their exhibiting efficient and high-quality generative capabilities in around ten scale steps, VAR models can suffer from diversity collapse under multiple samplings (see \cref{fig:teaser}) like few-step diffusion models. 
% This problem has also been observed in previous diffusion models when distilled into few-step versions~\cite{luo2023latent,sauer2023adversarial,lin2024sdxllightning,dao2024swiftbrush}.
However, although VAR models perform inference with far fewer steps than traditional AR models, they suffer from the \textit{diversity collapse} problem under numerous scenarios (\cref{fig:teaser}, 1st, 3rd and 5th rows). 
% , similar to few-step diffusion models.
% However, this often suffers from diversity collapse under multiple samplings~\cite{kang2024distilling, li2025one}.

% To address this challenge, xxx.
In this work, we address the \textit{diversity collapse} problem in VAR models by unlocking their inherent generative diversity. Through a detailed analysis of the coarse-to-fine next-scale prediction process in pretrained VAR models, we arrive at the following key observations.
\textit{\textbf{(a) Structural Formation in Early Scales:}} 
%The combined evidence in \cref{fig:Observation_1,fig:Observation_1_2} demonstrates that structural formation is initiated at early scales, while later scales primarily show stabilized structures. Therefore, our focus is directed toward the early scales.
As shown in \cref{fig:Observation_1,fig:Observation_1_2}, structural formation predominantly occurs at early scales, while later scales mainly refine and stabilize existing structures. This finding directs our attention toward the early-stage prediction dynamics.
\textit{\textbf{(b) Diversity Governed by the Pivotal Token:}} 
As demonstrated in \cref{fig:Observation_2,fig:Observation_2_2}, the pivotal token dominates the formation of structure while the auxiliary token carries both semantic information and image fidelity.
% the pivotal token is primarily responsible for structural formation, while the auxiliary token carries both semantic information and image fidelity.

Based on the these insights, we propose \textbf{DiverseVAR}, a training-free framework that effectively unleashes the inherent generative diversity of VAR models. 
%Rather than retraining the model, our method intervenes the inference process guided by our observations.
Without any additional training, DiverseVAR strategically intervenes during the inference process, guided by our analytical findings.
%It consists of two main steps. 
Our approach consists of two complementary sampling steps:
\textit{\textbf{(a) Soft-Suppression Regularization:}} We identify the pivotal component of the model input that governs diversity formation and apply a soft-suppression strategy to attenuate components contributing to structural redundancy, thereby mitigating diversity collapse.
\textit{\textbf{(b) Soft-Amplification Regularization:}} To further promote controlled diversity, we enhance the pivotal component of the model output through a soft-amplification mechanism, ensuring that the expanded diversity remains consistent text–image alignments.
%In the first step, we aim to unleash the generative diversity of VAR models. To achieve this, we first identify the pivotal component of the model input that is closely related to diversity formation. We then introduce a \textit{soft-suppression regularization} strategy that explicitly suppresses the components contributing to structural similarity, which otherwise lead to diversity collapse.
%In the second step, to further improve and guide the formation of diversity, we propose a \textit{soft-augmentation regularization} that augments the pivotal component of the model output to prevent unconstrained diversity from compromising text–image alignment.
%Together, these complementary methods ensure that the inherent generative diversity of VAR models is effectively unleashed, while maintaining fidelity in the generated results.
Together, these two regularizations enable \textit{DiverseVAR} to unleash the inherent generative potential of VAR models while preserving high image fidelity and faithful semantic alignment.
Our main contributions are summarized as follows:
\begin{itemize}[leftmargin=*]
% \begin{itemize}
    \item We conduct a systematic analysis to identify the scale factors that govern the \textit{diversity collapse} problem in VAR models. Our findings reveal that the VAR model diversity is predominantly influenced by the pivotal component at early scales.
    \item Based on our observations, we introduce \textbf{DiverseVAR}, a simple yet effective training-free method to enable the emergence of the inherent generative diversity of VAR models. It primarily incorporates two regularization terms during VAR inference: soft-suppression regularization and soft-amplification regularization.
    \item Experiments on the COCO, GenEval, and DPG benchmarks show that our method \textbf{DiverseVAR} unleashes the VAR model diversity while maintaining the sampled image fidelity, with slight influences on the generation performance according to numerous evaluation metrics.%, e.g. 0.03 reduction on GenEval.
\end{itemize}

\section{Related Wrok}
\subsection{Visual Autoregressive Generation}
Traditional autoregressive (AR) methods~\cite{lee2022autoregressive, sun2024llamagen, yu2022scaling} follow the next-token prediction paradigm, which requires a large number of iterative steps to generate high-quality images. Recently, Visual Autoregressive (VAR) modeling~\cite{tian2024visual} has adopted a next-scale prediction paradigm, enabling progressive generation across different resolutions to produce high-quality images. In contrast to AR methods that require massive steps to produce a high-resolution image~\cite{lee2022autoregressive, sun2024llamagen, yu2022scaling}, VAR can achieve comparable quality within only around ten steps~\cite{tian2024visual, Infinity, ma2024star, tang2024hart}.
Despite the promising performance of VAR models, current VAR models remain exhibit a notable gap: their generated samples for a given text prompt tend to be highly similar. Such indicated limited diversity is termed as \textit{diversity collapse} problem in this paper.

\begin{figure*}
    \centering
    \includegraphics[width=\linewidth]{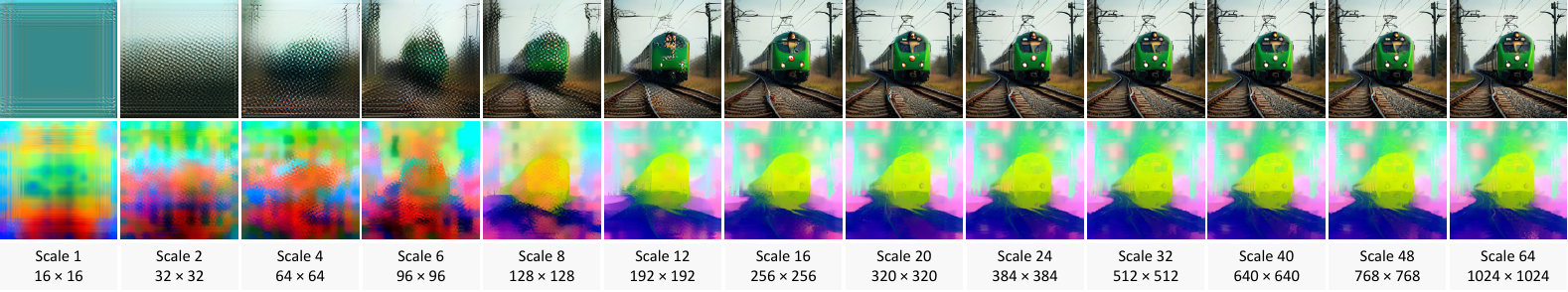}\vspace{-2mm}
    \caption{Visualization of samples across all scales (1st row) and their associated DINO features (2nd row).}\vspace{-4mm}
    \label{fig:Observation_1}
\end{figure*}
% prompt: a green train is coming down the tracks

\begin{figure}
    \centering
    \includegraphics[width=\linewidth]{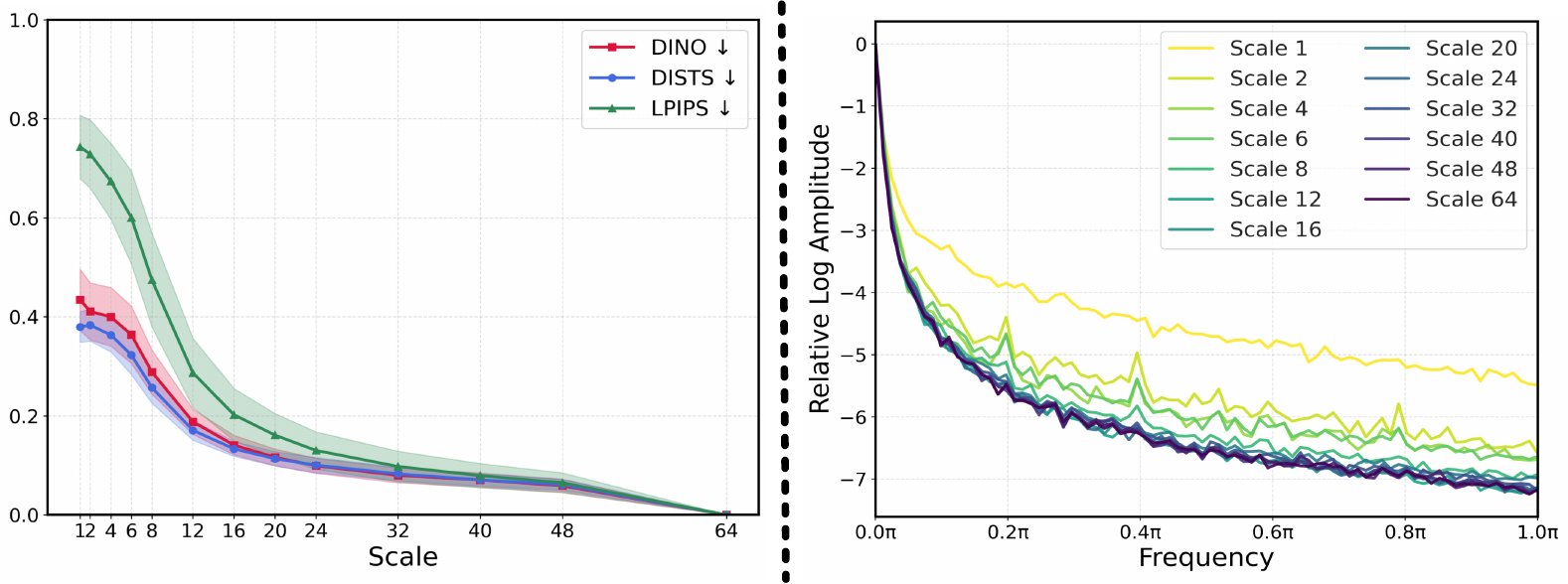}\vspace{-2mm}
    \caption{(Left) Statistics of structure evolution on all scale steps. (Right) The relative log amplitude of frequency components across different scales.}\vspace{-4mm}
    \label{fig:Observation_1_2}
\end{figure}

% \subsection{Diverse Visual Generation}
\subsection{Alternative Visual Generation Paradigms}
\textbf{Diffusion models}~\cite{rombach2022high,podell2023sdxl} trained on large-scale text-image datasets~\cite{schuhmann2022laion} are able to generate high-quality and diverse images, but they face a time-consuming challenge due to multi-step sampling.
To address this challenge, distillation techniques~\cite{luo2023latent,sauer2023adversarial,lin2024sdxllightning,dao2024swiftbrush} have been applied to diffusion models, enabling few-step sampling in distilled student models. Despite the improved sampling speed, the resulting images often exhibit \textit{diversity collapse}~\cite{kang2024distilling}.
The exploration of diverse generation has recently attracted increasing attention in the context of few-step diffusion models. For example, Diffusion2GAN~\cite{kang2024distilling} aligns its generation trajectory with that of the teacher diffusion model, effectively reducing diversity collapse. Loopfree~\cite{li2025one} employs a 4-step parallel decoder in the UNet architecture combined with Variational Score Distillation (VSD)~\cite{wang2023prolificdreamer} to generate images that exhibit both high diversity and fidelity. Hybrid~\cite{gandikota2025distilling} uses the base model only for the first step to seed diversity, and then switches to the distilled model for subsequent generation. C3~\cite{han2025enhancing} enhances generative diversity by amplifying internal feature representations using automatically selected amplification factors.
Despite improving diversity, the requirement for training remains a bottleneck~\cite{kang2024distilling,li2025one}, the simultaneous use of both teacher and student models further increases memory consumption~\cite{gandikota2025distilling}, and depend on creativity-oriented prompts to stimulate diverse outputs~\cite{han2025enhancing}.
By contrast, \textbf{AR models} often demonstrate strong generative diversity~\cite{xiong2024autoregressive}. Recent researches~\cite{ma2025betterfasterautoregressive,yu2024randomized} focus on enhancing the balance between content quality and diversity, rather than specifically addressing diversity in AR models.
In contrast to conventional AR models, \textit{VAR} performs next-scale prediction to enable efficient generation, but suffers from diversity collapse. Nevertheless, prior efforts addressing diversity in diffusion and AR models are not directly transferable to VAR, and exploration of diversity in VAR remains scarce.

In this work, we conduct a systematic analysis to identify which scales govern diversity in VAR models. Building on these observations, we further characterize the properties of these scales that are closely associated with diversity. Finally, leveraging these properties, we propose a diversity enhancement technique for VAR inference that preserves both semantic consistency and visual fidelity.
% To the best of our knowledge, this study is the first to identify and address the diversity challenge in VAR models.
% \input{sec/2_formatting}
%% TongWang：图片两种排放方式，选一种
% \begin{figure*}
%     \centering
%     \includegraphics[width=\linewidth]{figure/figure2_pre_study1_compressed.pdf}\vspace{-2mm}
%     \caption{Visualization of samples across all scales (1st row) and their associated DINO features (2nd row).}\vspace{-4mm}
%     \label{fig:Observation_1}
% \end{figure*}

% \begin{figure}
%     \centering
%     \includegraphics[width=\linewidth]{figure/figure2_pre_study1_2_compressed.pdf}\vspace{-2mm}
%     \caption{(Left) Visualization of structure evolution on all scale steps. (Right) The relative log amplitude of frequency components across different scales.}\vspace{-4mm}
%     \label{fig:Observation_1_2}
% \end{figure}

\begin{figure*}
    \centering
    \includegraphics[width=\linewidth]{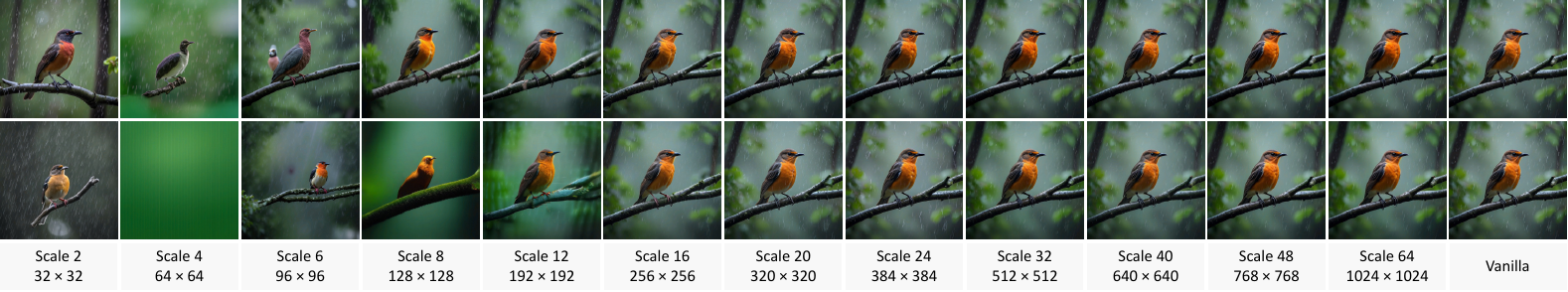}\vspace{-2mm}
    \caption{Visualization of samples when zeroing out the pivotal (1st row) or auxiliary (2nd row) tokens across all scales except the 1st scale (1st–12th columns), along with the vanilla generation results (last column).}\vspace{-4mm}
    \label{fig:Observation_2}
\end{figure*}
% prompt: a bird perched on a tree branch in the rain

\begin{figure}
    \centering
    \includegraphics[width=\linewidth]{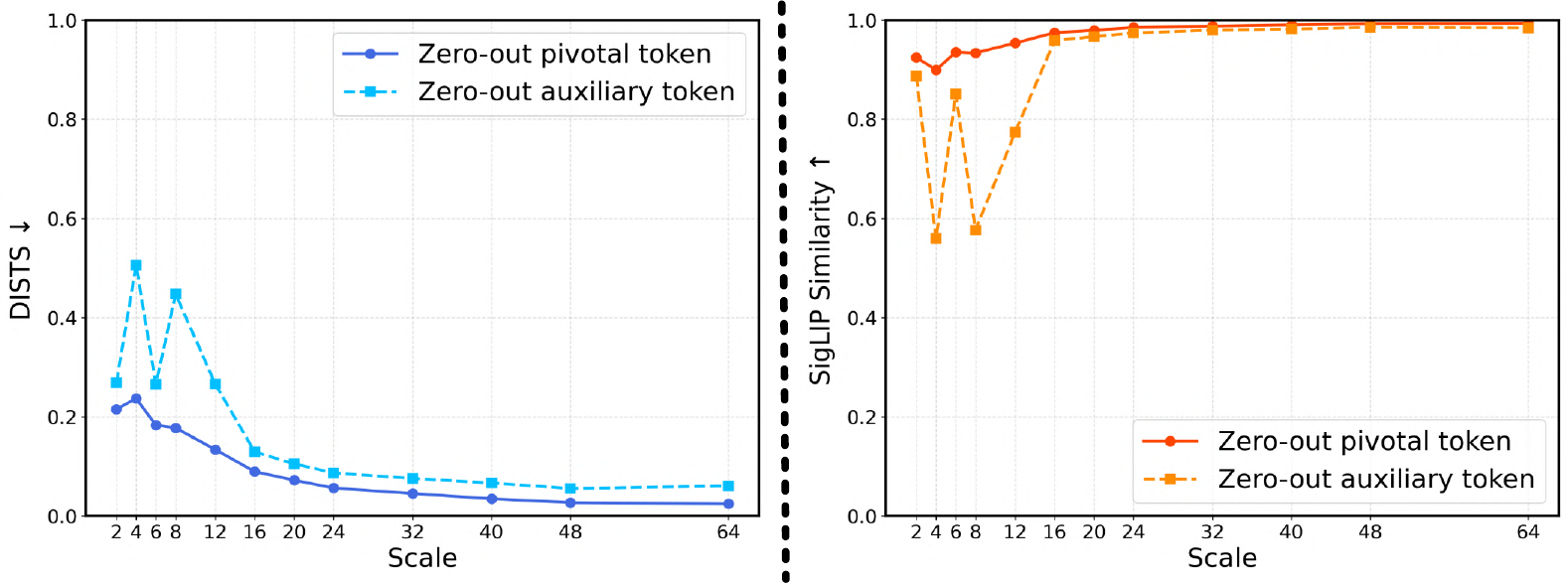}\vspace{-2mm}
    \caption{Structural (Left) and semantic (Right) evaluation when pivotal and auxiliary tokens are zeroed out.}\vspace{-4mm}
    \label{fig:Observation_2_2}
\end{figure}

\section{Method}
VAR revisits the conventional next-token prediction paradigm in AR image generation and introduces a coarse-to-fine next-scale prediction framework (\cref{subsec:prelim}), which substantially accelerates sampling while generating high-quality images, yet suffers from \textit{diversity collapse}.
In this section, we explore how the mechanisms underlying the coarse-to-fine generation process affect diversity (\cref{subsec:motivation}). Building on these insights, we propose our method \textit{DiverseVAR} that fosters the emergence of diversity, enabling VAR models to achieve higher generative diversity while maintaining generation fidelity (\cref{subsec:ours}).

\subsection{Preliminaries}\label{subsec:prelim}
Visual Autoregressive (VAR) modeling~\citep{tian2024visual} reformulates the conventional autoregressive (AR) framework~\citep{lee2022autoregressive,razavi2019generating,sun2024llamagen,yu2022scaling} to the visual domain by transitioning from next-token to next-scale prediction. Under this paradigm, each autoregressive step produces a token map corresponding to a specific resolution scale, instead of predicting tokens one by one.
Given an image $\boldsymbol{I}\in\mathbb{R}^{H\times W\times3}$, a continuous image feature map $\boldsymbol{F}\in\mathbb{R}^{h\times w\times d}$ is obtained through a visual tokenizer. VAR then quantizes this feature map into $K$ multi-scale token maps $\boldsymbol{R}=(\boldsymbol{R}_1, \boldsymbol{R}_2, \ldots, \boldsymbol{R}_K)$, where each token map corresponds to a predefined spatial resolution $(h_k, w_k)$ for $k=1,\ldots,K$. Here, $(h_K, w_K)$ denotes the final scale, which is identical to the spatial size $(h, w)$ of the feature map $\boldsymbol{F}$.
This sequence of residuals progressively approximates the continuous feature map $\boldsymbol{F}$ as:
\begin{equation}
\boldsymbol{F}_k = \sum_{i=1}^k \mathrm{up}(\boldsymbol{R}_i, (h,w)),
\label{eq:var_upf}
\end{equation}
where $\mathrm{up}(\cdot)$ denotes the upsampling operation. The transformer autoregressively predicts the next-scale 
% residual 
$\boldsymbol{R}$ conditioned on the previously generated residuals, with the overall likelihood formulated as: 
\begin{equation}%\vspace{-1mm}
p(\boldsymbol{R}_1, \ldots, \boldsymbol{R}_K) = \prod_{k=1}^K p(\boldsymbol{R}_k \mid \boldsymbol{R}_1, \ldots, \boldsymbol{R}_{k-1}). 
\label{eq:var_joint_p}%\vspace{-1mm}
\end{equation}

The transformer block in VAR takes the text embeddings at the first scale as input to predict $\boldsymbol{R}_1$. In the subsequent $k$-th scale, the feature map from the previous $(k\text{-}1)$-th scale is downsampled to match the spatial resolution of the input at the $k$-th scale:
\begin{equation} 
\widetilde{\boldsymbol{F}}_{k-1} = \mathrm{down}(\boldsymbol{F}_{k-1},(h_k,w_k)),
\label{eq:var_downf}
\end{equation}
where $\mathrm{down}(\cdot)$ denotes the downsampling operation, and the transformer block receives the downsampled feature $\widetilde{\boldsymbol{F}}_{k-1}$ as input to predict $\boldsymbol{R}_k$.

\subsection{Motivation}\label{subsec:motivation}
In this subsection, we present the motivation for our proposed method, based on two key observations\footnote{The observations are drawn by statistical experiments on generated images, using 100 random text prompts from the COCO dataset~\cite{lin2014microsoft}.} derived from vanilla text-to-image generative models~\cite{Infinity}.
First, the structural formation in vanilla next-scale models occurs at early scales, motivating us to explore the mechanisms that influence model diversity during these early scales (\cref{fig:Observation_1,fig:Observation_1_2}). Second, model diversity is primarily influenced by the pivotal token at early scales, presenting an opportunity to further promote the emergence of diversity while preserving generation fidelity (\cref{fig:Observation_2,fig:Observation_2_2}).

\minisection{\textit{Observation 1: Structural Formation in Early Scales.}}
We experimentally observe that the structural formation emerges at early scales, whereas later scales exhibit already stabilized structures (\cref{fig:Observation_1,fig:Observation_1_2}). 
More specifically, given a pretrained VAR model, it progressively produces intermediate predictions $\boldsymbol{R}_k$ and constructs the corresponding feature maps $\boldsymbol{F}_k$ (\cref{eq:var_upf}) as the scale gradually increases (\cref{fig:Observation_1} (1st row)). 
% As illustrated in~\cref{fig:Observation_1} (1st row), .

Motivated by~\cite{tumanyan2022splicing,Tumanyan_2023_CVPR,park2024energy}, we adopt DINO features~\cite{oquab2023dinov2} as a representation of the structural characteristics of the feature maps (\cref{fig:Observation_1} (2nd row)).
As illustrated in~\cref{fig:Observation_1} (2nd row), the structural formation of the image persists up to a specific scale (i.e., 12), after which the structure stabilizes.
Here, we further utilize DINO structure distance~\cite{tumanyan2022splicing,Tumanyan_2023_CVPR} to measure whether the feature maps exhibit a well-formed overall structure comparable to that of the final scale (\cref{fig:Observation_1_2}). As illustrated in~\cref{fig:Observation_1_2} (Left), the curve quickly fall below 0.2 at early scales (i.e., 12), indicating ongoing structural formation. The curve then converge at the remaining scales, implying that the structure has already stabilized.
We also employ LPIPS~\cite{zhang2018unreasonable} and DISTS~\cite{ding2020iqa} to evaluate structural formation. \cref{fig:Observation_1_2} (Left) shows that the curves of both LPIPS and DISTS exhibit a similar trend to that of the DINO structure distance.
Furthermore, we perform frequency-domain analysis of the DINO features to evaluate the progression of structural formation (\cref{fig:Observation_1_2} (Right)). The frequency components decrease sharply at the early scales and gradually converge at the later scales, indicating that they become more stable and consistent as the scale increases. 
In conclusion, the early scales can be exploited to modify the structure, thereby promoting the emergence of diversity.

% \textbf{Diversity Governed by Principal Components.}
\minisection{\textit{Observation 2: Diversity Governed by Pivotal Token.}}
% From the above observation and analysis, we infer that the structural formation primarily emerges at early scales. 
The above observation and analysis indicate that structural formation primarily occurs at early scales. This finding motivates us to further examine the components that influence this formation.
%
% Motivated by~\cite{guo2025fastvar}, we .
% Following~\cite{guo2025fastvar}, 
Thus, we investigate the internal composition of the feature map $\widetilde{\boldsymbol{F}}_{k-1}$ at scale $k$ (\cref{eq:var_downf}) by dividing it into \textit{pivotal} and \textit{auxiliary} components.
% in order to elucidate their respective roles in the structural formation process.
Specifically, we follow~\cite{guo2025fastvar} and define the pivotal score $s_{k,i} = \|\widetilde{\boldsymbol{F}}_{k-1,i}-\bar{\boldsymbol{F}}_{k-1}\|_2$ using the L2 norm,
% \begin{equation} 
% s_k = \|\boldsymbol{F}_k-\bar{\boldsymbol{F}}_k\|_2,
% \label{eq:sk}
% \end{equation}
where $\bar{\boldsymbol{F}}_{k-1}$ represents the mean feature map obtained by averaging $\widetilde{\boldsymbol{F}}_{k-1}$ across scale dimensions, and $\widetilde{\boldsymbol{F}}_{k-1,i}$ denotes each token along the scale dimension.
Based on their pivotal scores, all tokens in the feature map are ranked and then categorized into \textit{pivotal} and \textit{auxiliary} tokens, with the partition determined by the elbow point computed using {the Maximum Distance to Chord (MDC) method~\cite{douglas1973algorithms}} (See \textit{\textcolor{blue}{Suppl.}} for details). We simply regard the $\textit{pivotal}$ and $\textit{auxiliary}$ tokens as the $\textit{pivotal}$ and $\textit{auxiliary}$ components, and refer to this straightforward partition approach as $\textit{Naive Component Partition}$ (NCP).
% as follows
% \begin{equation} 
% \boldsymbol{F}_k = \left\{\begin{array}{ll}
% \sigma \lambda_{i}, & \text { if } i \leq k \\
% \lambda_{i}, & \text { if } i>k,
% \end{array}\right.
% \label{eq:var_downf}
% \end{equation}
% .

We observe that the pivotal token primarily contribute structural formation, whereas the auxiliary token carry both semantic information and image fidelity.
For example, when provided with the text prompt ``A bird perched on a tree branch in the rain'', VAR generates an image consistent with the given description (\cref{fig:Observation_2} (last column)).
% When zeroing out the pivotal tokens at a specific scale (e.g., 4), the generated image exhibits structural changes while preserving semantic content.
When zeroing out the pivotal token at one of the early scales (e.g., scale 4 within scales 2–8), the generated image exhibits noticeable structural changes, while its semantic meaning remains consistent (\cref{fig:Observation_2} (1st row, 1st-4th columns).
In contrast, zeroing out the auxiliary token at one of these early scales severely degrades both semantic information and image fidelity (\cref{fig:Observation_2} (2nd row, 1st-4th columns).
Furthermore, we employ DISTS~\cite{ding2020iqa} and SigLIP~\cite{tschannen2025siglip} to quantitatively evaluate the modified images in comparison with the vanilla VAR outputs.
As shown in~\cref{fig:Observation_2_2}, DISTS and SigLIP display consistent trends at early scales: zeroing out pivotal token (\underline{solid lines}) results in slight degradation, while zeroing out auxiliary token (\underdashline{dashed lines}) results in a drastic deterioration in both metrics. 
For example, when zeroing out the pivotal token, the variances of DISTS and SigLIP are below 0.3 and 0.1, respectively, whereas the corresponding values for the auxiliary token exceed 0.5 and 0.4.
Finally, when zeroing out either the pivotal or auxiliary token at one of the later scales (e.g., scales 16-64), both the structural and semantic characteristics remain consistent with those of the vanilla generation (\cref{fig:Observation_2} (6th-12th columns) and \cref{fig:Observation_2_2}).
These results indicate that the model’s generative diversity is predominantly governed by the pivotal token at early scales.

However, using the \textit{Naive Component Partition} (NCP) to identify the pivotal component and apply its suppression (i.e., zeroing out) can promote the emergence of generative diversity, yet this naive approach may also induce unexpected variations (\cref{fig:Observation_2}, 1st row, 2nd and 3rd columns).

\begin{figure}
    \centering
    \includegraphics[width=\linewidth]{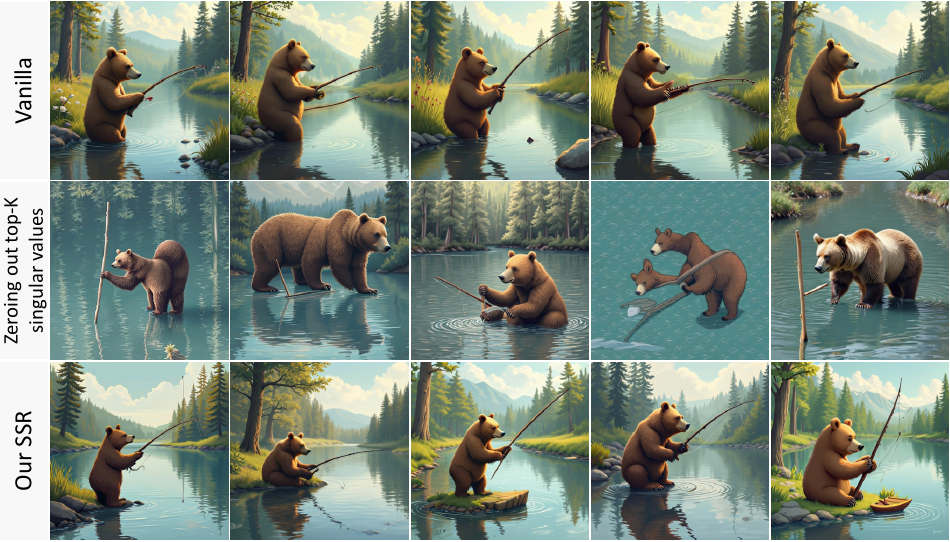}\vspace{-2mm}
    \caption{The dominant singular values correspond to the pivotal component influence generative diversity}\vspace{-4mm}
    \label{fig:zeroing_s0}
\end{figure}
% prompt: A bear fishing with a stick by a calm river.

% 合并figure
\begin{figure}
    \centering
    \includegraphics[width=1\linewidth]{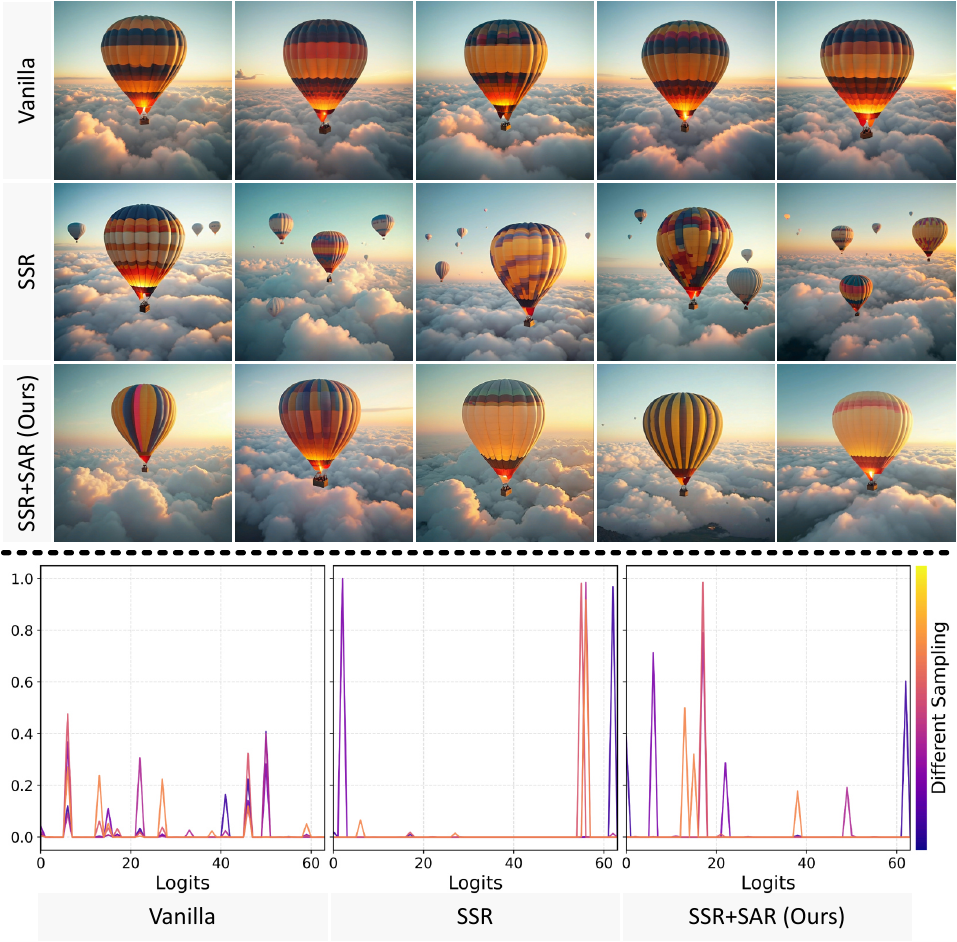}\vspace{-2mm}
    \caption{(Top) The generated images often fail to reflect the number described in the text prompt when using only SSR. For example, given the prompt “\textbf{A hot air balloon} floating above the clouds”, SSR fails to generate the correct quantity of the \textbf{ hot air balloon} (2nd row). (Bottom) The logits distribution under different samplings of vanilla model (Left), SSR (Middle), and SSR+SAR (Right).}\vspace{-4mm}
    \label{fig:step2}
\end{figure}

\subsection{Diverse Visual Generation for VAR}\label{subsec:ours}
\begin{figure*}
    \centering
    \includegraphics[width=\linewidth]{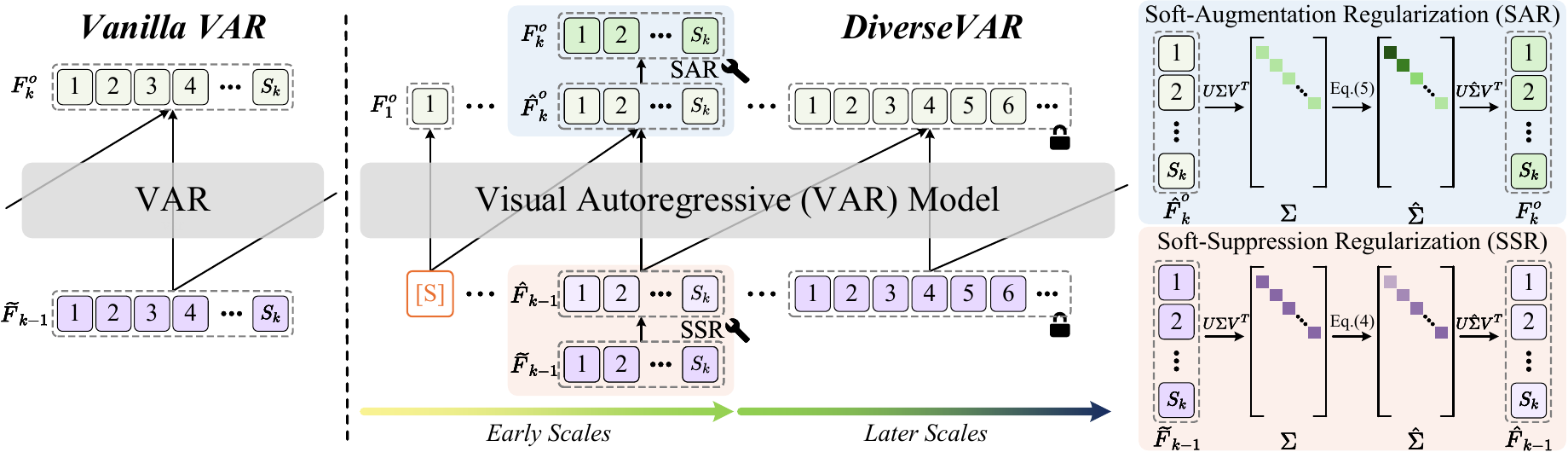}\vspace{-2mm}
    \caption{The overall framework of \textbf{DiverseVAR}. We explore the diversity emergence of VAR models at the early scales, while retaining the original VAR inference process at the later scales.}\vspace{-4mm}
    \label{fig:framework}
\end{figure*}

Based on all the above observations, we propose \textbf{DiverseVAR} for VAR to trigger the emergence of generative diversity while preserving generation fidelity in the VAR inference (\cref{fig:framework} and \cref{alg:diversevar} (See \textit{\textcolor{blue}{Suppl.}})).
As shown in~\cref{fig:framework} (Left), the vanilla VAR inference processes the intermediate feature $\widetilde{\boldsymbol{F}}_{k-1}$ through a VAR block to produce the output feature $\boldsymbol{F}_k^o$ as well as
its predicted logits, which is subsequently quantized into $\boldsymbol{R}_k$ (The predicted logits, the quantization step, and $\boldsymbol{R}_k$ are omitted in~\cref{fig:framework} for brevity).
Our method operates at the block level (e.g., 8 blocks in the Infinity backbone).

\minisection{Diversity emergence via Pivotal Component.}
For the pivotal component, it is difficult to disentangle it within the pivotal token, as the self-attention mechanism in Transformer-based VAR models inherently entangles information across tokens~\cite{vaswani2017attention,Infinity} in the feature map $\widetilde{\boldsymbol{F}}_{k-1}$.
Inspired by~\cite{gu2014weighted}, we assume that the dominant singular values of the feature map $\widetilde{\boldsymbol{F}}_{k-1}$ correspond to the fundamental information, i.e., the pivotal component. To avoid directly using the pivotal token as the pivotal component, we instead regard the dominant singular values as determining the pivotal component.
Specifically, given the feature map $\widetilde{\boldsymbol{F}}_{k-1}\in \mathbb{R}^{S_k \times D}$ at the $k$-th (early) scale~\footnote{i.e., $S_k=h_k\times w_k$, $D=2048$ in the Infinity model~\citep{Infinity}}, we decompose it using Singular Value Decomposition (SVD): $\widetilde{\boldsymbol{F}}_{k-1}=\mathbf{U}{\mathbf\Sigma}{\mathbf{V}^T}$, where $\mathbf\Sigma= diag(\sigma_1, \cdots, \sigma_{n})$, the singular values $\sigma_1 \geq \cdots \geq \sigma_n$, $n={\rm min}(S_k,D)$. 
Then, our objective is to suppress the pivotal component, thereby triggering the emergence of generative diversity. 
To suppress the pivotal component of the feature map $\widetilde{\boldsymbol{F}}_{k-1}$, we introduce the \textit{Soft-Suppression Regularization} (\textbf{SSR}) for each singular value, which is formulated as:
\begin{equation}
\hat{\sigma} = \alpha{e}^{-\beta\sigma}\times\sigma.
\label{eq:ssr}
\end{equation}
Here, $e$ represents the exponential function, while $\alpha$ and $\beta$ are parameters constrained to be positive. For brevity, the subscript of $\sigma$ is omitted.
% We then recover the feature map $\hat{\boldsymbol{F}}_{k-1}=\mathbf{U}{\hat{\mathbf\Sigma}}{\mathbf{V}^T}$ with the updated $\hat{\mathbf\Sigma}= diag(\hat{\sigma}_1, \cdots, \hat{\sigma}_{n})$
The feature map is then reconstructed as $\hat{\boldsymbol{F}}_{k-1} = \mathbf{U} \hat{\mathbf{\Sigma}} \mathbf{V}^T$, with $\hat{\mathbf{\Sigma}} = \operatorname{diag}(\hat{\sigma}_1, \dots, \hat{\sigma}_n)$ denoting the updated singular values. The feature map $\hat{\boldsymbol{F}}_{k-1}$ is passed through a block to generate the output $\hat{\boldsymbol{F}}_k^o$.

In a special case, we reset the top-K singular values to 0 (here, K=2). Interestingly, as shown in~\cref{fig:zeroing_s0} (2nd row), removing the dominant components leads to improved generative diversity but reduces image fidelity. This supports our assumption that the dominant singular values of the feature map correspond to the pivotal component that influence generative diversity (\cref{fig:zeroing_s0} (2nd row)), and also shows that our method suppresses the pivotal component more effectively than directly zeroing it out (\cref{fig:zeroing_s0} (3rd row)).

% \begin{figure}
%     \centering
%     \includegraphics[width=\linewidth]{figure/figure6_logits_comparison.pdf}\vspace{-2mm}
%     \caption{The logits distribution under different samplings of vanilla model (Left), SSR (Middle), and SSR+SAR (Right).}\vspace{-2mm}
%     \label{fig:logits_comparison}
% \end{figure}

% \begin{figure}
%     \centering
%     \includegraphics[width=1\linewidth]{figure/figure9_ssr_number_compressed.pdf}
%     % A hot air balloon floating above the clouds.
%     \caption{\textbf{todo: ssr counting problem}}
%     \label{fig:ssr_number}
% \end{figure}

% % 合并figure
% \begin{figure}
%     \centering
%     \includegraphics[width=1\linewidth]{figure/figure10_ssr_logits_revised_compressed.pdf}
%     \caption{\textbf{todo}}
%     \label{fig:step2}
% \end{figure}

% \begin{table}[tb]
% \renewcommand{\arraystretch}{1} % 控制行距（行高）
% \tabcolsep=0.02cm % 控制列间距
% \centering
% \resizebox{\linewidth}{!}{
% \begin{tabular}{c|ccccccc}
% \toprule
% {Dataset} 
% & \multicolumn{7}{c}{{GenEval}}  \\
% \cmidrule{1-1}\cmidrule{2-8}
% Method &
% Two Obj. & Pos. & Color Attri. & Count. & Colors & Single Obj.& {{Overall$\uparrow$}}  \\
% \midrule
% Vanilla & 0.84 & 0.45 & 0.55 & 0.70 & 0.84 & 1.00 & 0.73 \\
% SSR & 0.81 & 0.39 & 0.53 & 0.53 & 0.82 & 0.99 & 0.68 \\
% SSR+SAR & 0.85 & 0.41 & 0.53 & 0.58 & 0.82 & 0.99 & 0.70 \\
% \bottomrule
% \end{tabular}
% }
% \vspace{-3mm}
% \caption{SSR+SAR. 
% }
% \vspace{-2mm}
% \label{tab:ssr_sar}
% \end{table}

\minisection{Guided Diversity Formation.}
\textit{Soft-Suppression Regularization} facilitates diversity emergence via the pivotal component. 
% However, we observe that it can also affect the representation of numerical attributes, thereby leading to suboptimal image-text alignment 
However, while enhancing diversity, it weakens the alignment with the text semantics, especially in cases involving numerical descriptions (\cref{fig:step2} (Top (2nd row))). 
% reducing the alignment between the generated images and the text prompt. 

% To explore the reason behind this phenomenon, we evaluate the distribution of the predicted logits of the output feature $\boldsymbol{F}_k^o$.
% To understand the underlying cause of this phenomenon, we analyze the distribution of the predicted logits for the output feature $\boldsymbol{F}_k^o$.
Motivated by~\cite{ma2025betterfasterautoregressive}, we analyze the distribution of the predicted logits for the output feature ${\boldsymbol{F}}_k^o$ to understand the underlying cause of this phenomenon.
We observe that the logits distributions under different samplings are similar in the vanilla VAR (\cref{fig:step2} (Bottom (Left))), while \textit{Soft-Suppression Regularization} leads to more dispersed logits distributions across different samplings (\cref{fig:step2} (Bottom (Middle))).
% Specifically, the logits distributions observed in the vanilla VAR reveal that the probability peaks of different samplings nearly coincide, leading to highly similar samples and a lack of diversity in the generated images. In contrast, incorporating \textit{Soft-Suppression Regularization} results in more dispersed logits distributions, where the peaks vary significantly in position and magnitude, thereby promoting diversity in the generated outputs.
Specifically, the logits distributions observed in the vanilla VAR reveal that the probability peaks of different samplings nearly coincide (\cref{fig:step2} (Bottom (Left))), leading to highly similar samples and a lack of diversity in the generated images. In contrast, incorporating \textit{Soft-Suppression Regularization} yields more dispersed logits distributions (\cref{fig:step2} (Bottom (Middle))), with peaks varying significantly in both position and magnitude, thereby promoting the emergence of diversity in the generated images.
However, we observe that independent and very high probability peaks can distort the representation of numerical attributes, leading to suboptimal image-text alignment.

% Here, we aim to further address inaccuracies in numerical attributes to improve image-text alignment while maintaining generative diversity. 
Here, we aim to further improve image-text alignment, especially reducing inaccuracies in numerical attributes, while maintaining generative diversity.
Based on the above analysis, we introduce an augmentation for the output feature ${\hat{\boldsymbol{F}}}_k^o$ to further guide the formation of diversity. Specifically, we perform singular value decomposition (SVD) on $\hat{\boldsymbol{F}}_k^o$, yielding singular values $\hat\sigma_1 \geq \cdots \geq \hat\sigma_n$.
We then augment each singular value according to the following rule, referred to as \textit{Soft-Amplification Regularization} (\textbf{SAR}):
\begin{equation}
\tilde{\sigma} = \hat\alpha{e}^{\hat\beta\hat\sigma}\times\hat\sigma,
\label{eq:sar}
\end{equation}
where $\hat\alpha$ and $\hat\beta$ are positive parameters controlling the scaling strength.
The recovered feature ${\boldsymbol{F}}_k^o=\mathbf{U}{\widetilde{\mathbf\Sigma}}{\mathbf{V}^T}$, where $\widetilde{\mathbf\Sigma}= diag(\tilde\sigma_1, \cdots, \tilde\sigma_{n})$. The \textit{Soft-Augmentation Regularization}, by encouraging more dispersed logit distributions and avoiding the formation of isolated peaks across different samplings, can further guide the formation of diversity (\cref{fig:step2} (Top (3rd row) and Bottom (Right))).
\section{Experiment}

\subsection{Experimental Setup}
\minisection{Evaluation Datasets and Metrics.}
We evaluate \textbf{DiverseVAR} on the text-to-image VAR models Infinity-2B and Infinity-8B~\cite{Infinity}, generating images at a resolution of 1024$\times$1024.
% Following the vanilla text-to-image Infinity model~\cite{Infinity}, we use the GenEval~\citep{ghosh2023geneval} and DPG~\citep{DPG-bench} benchmarks to assess semantic alignment and perceptual quality of generated images~\citep{Infinity,li2025scalekv,chen2025sparsevar}. 
Following the setup of the vanilla Infinity-2B and Infinity-8B models~\cite{Infinity}, we evaluate \textbf{DiverseVAR} on two widely used benchmarks~\citep{Infinity,li2025scalekv,chen2025sparsevar}: GenEval~\citep{ghosh2023geneval} and DPG~\citep{DPG-bench}.
% showing that our method improves generative diversity without compromising semantic alignment or perceptual quality.
We further evaluate the generative diversity of our method in comparison with the vanilla model on the zero-shot text-to-image benchmarks COCO 2014 and COCO 2017~\cite{lin2014microsoft}.
On the COCO 2014 benchmark, we adopt the conventional evaluation protocol~\cite{sauer2023stylegant, kang2023scaling, saharia2022photorealistic, rombach2022high}, using the 30K text prompts selecte by GigaGAN~\cite{kang2023gigagan} to generate 30K corresponding images.
On the COCO 2017 benchmark, we generate 5K images from the 5K provided text prompts.
Following prior works~\cite{kang2023StudioGANpami,han2025enhancing}, we evaluate the generative diversity of the synthesized images using Fréchet Inception Distance (FID)~\cite{heusel2017gans}, Recall~\cite{Kynkaanniemi2019}, and Coverage (Cov.)~\cite{naeem2020reliable}. To assess the text-image alignment, we employ CLIPScore (CLIP)~\cite{hessel2021clipscore}, where ViT-B/32 is used as the backbone for feature extraction.

\minisection{Implementation Details.}
Following our observations that diversity formation primarily occurs at early scales, we apply our method at scales $\{4, 6\}$ to unleash diversity, while retaining the vanilla inference process in the remaining scales $\{1,2,8,12,16,20,24,32,40,48,64\}$ to maintain fidelity.
Our operation is applied at the block level, spanning all 8 blocks of the Infinity backbone.
We set the parameters as $\alpha=1.0$ and $\beta=0.01$ in~\cref{eq:ssr}, and $\hat{\alpha}=1.0$ and $\hat{\beta}=0.001$ in~\cref{eq:sar}.
All experiments are performed on a single NVIDIA A100 GPU equipped with 40 GB of memory.
See \textit{\textcolor{blue}{Suppl.}} for details.

\subsection{Main Results}
\begin{figure}
    \centering
    \includegraphics[width=\linewidth]{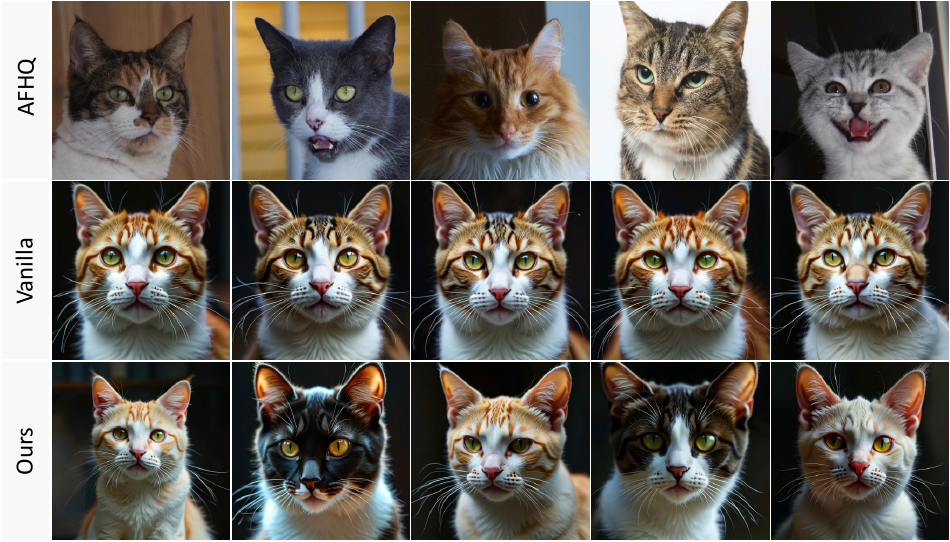}
    \caption{The vanilla model tends to produce results with similar styles under the same prompt, leading to limited diversity, while our results are more consistent with those from AFHQ.}\vspace{-2mm}
    \label{fig:afhq_cat}\vspace{-2mm}
\end{figure}

\minisection{Diversity Comparison.}
To demonstrate the generative diversity of our \textbf{DiverseVAR}, we report quantitative evaluations using Recall, Coverage, and FID. 
% Generally, Recall represents the ratio of real images contained within the support of the generated data. Coverage measures the extent to which the generated images cover all significant regions of the real data distribution.
Generally, Recall and Coverage are commonly used to assess the diversity of generated samples by comparing the support of real and generated data distributions~\cite{kang2023StudioGANpami,naeem2020reliable}.
% FID measures the distributional similarity between real and generated datasets.
% As reported in~\cref{tab:fid_clip}, our method achieves higher Recall, Coverage, and FID than the vanilla Infinity on both benchmarks, demonstrating improved diversity while maintaining text-image alignment, as confirmed by CLIP.
As reported in~\cref{tab:fid_clip}, our method achieves higher Recall, Coverage, and FID than the vanilla Infinity on both benchmarks, while maintaining comparable CLIP scores.
\cref{fig:teaser} shows generations from the vanilla Infinity and our method, qualitatively demonstrating that our approach produces more diverse outputs.
Furthermore, to evaluate the diversity under multiple samplings, we use booth AFHQ~\cite{choi2020starganv2} and CelebA-HQ~\cite{karras2017progressive}.
We use three prompts corresponding to the three categories of AFHQ and two prompts corresponding to CelebA-HQ, each generating approximately 5,000 images. During evaluation, the training sets of both AFHQ and CelebA-HQ are used as the ground truth. 
As reported in~\cref{tab:fid_AFHQ}, compared with the vanilla Infinity-2B and Infinity-8B, our method achieves superior scores in terms of Recall, Coverage, and FID on both datasets, except for Recall on the Infinity-8B model. \cref{fig:afhq_cat} illustrates that our generated results (3rd row) more closely resemble the real images from AFHQ (1st row).
See \textit{\textcolor{blue}{Suppl.}} for details.

\begin{table}[tb]
\renewcommand{\arraystretch}{1} % 行高
\tabcolsep=0.03cm % 列宽
\centering
\resizebox{1\linewidth}{!}{
\begin{tabular}{c|ccc|c|ccc|c}
\toprule
{Dataset}
& \multicolumn{4}{c|}{{COCO2014-30K}} & \multicolumn{4}{c}{{COCO2017-5K}} \\
% \cmidrule{1-1}\cmidrule{5-18}
\midrule
Method & {{Recall$\uparrow$}} & {{Cov.$\uparrow$}} &
{{FID$\downarrow$}} & {{CLIP$\uparrow$}} &
{{Recall$\uparrow$}}  & {{Cov.$\uparrow$}} & 
{{FID$\downarrow$}} & {{CLIP$\uparrow$}}  \\
\midrule
% Infinity
Infinity-2B &
0.316  & 0.651 & 28.48 & 0.313 &
0.408  & 0.832 & 39.01 & 0.313 \\
% ours-stage2
\lightredrow\textbf{+Ours} &
\textbf{0.385}  & \textbf{0.690} & \textbf{22.96} & 0.313 & 
\textbf{0.480} & \textbf{0.860} & \textbf{33.39} & 0.313 \\
\bottomrule
% 8B
Infinity-8B &
0.451  & 0.740 & 18.79 & 0.319 &
0.563  & 0.892 & 29.47 & 0.319 \\
% 8B-ours
\lightredrow\textbf{+Ours} &
\textbf{0.497}  & \textbf{0.748 }& \textbf{14.26} & 0.315 & 
\textbf{0.585}  & \textbf{0.892} & \textbf{25.01} & 0.316 \\
\bottomrule
\end{tabular}
}
\vspace{-3mm}
\caption{Quantitative comparison between the vanilla model and our DiverseVAR on COCO2014 (30K prompts) and COCO2017 (5K prompts), evaluated using FID, Recall, and Coverage (Cov.) for generative diversity, and CLIPScore (CLIP) for text–image alignment.}
\vspace{-2mm}
\label{tab:fid_clip}\vspace{-2mm}
\end{table}

% 以下是 afhq 和 celeba_hq
\begin{table}[tb]
\renewcommand{\arraystretch}{1} % 行高
\tabcolsep=0.03cm % 列宽
\centering
\resizebox{1\linewidth}{!}{
\begin{tabular}{c|ccc|c|ccc|c}
\toprule
{Dataset}
& \multicolumn{4}{c|}{{AFHQ}} & \multicolumn{4}{c}{{CelebA-HQ}} \\
% \cmidrule{1-1}\cmidrule{5-18}
\midrule
Method & {{Recall$\uparrow$}} & {{Cov.$\uparrow$}} &
{{FID$\downarrow$}} & {{CLIP$\uparrow$}} &
{{Recall$\uparrow$}}  & {{Cov.$\uparrow$}} & 
{{FID$\downarrow$}} & {{CLIP$\uparrow$}}  \\
\midrule
% 2B
Infinity-2B &
0.0  & 0.007 & 91.85 & 0.276 &
0.003  & 0.007 & 81.16 & 0.259 \\
% 2B-ours
\lightredrow\textbf{+Ours} &
0.012  & 0.020 & 78.88 & 0.278 &
0.017  & 0.128 & 62.07 & 0.257 \\
\bottomrule
% 8B
Infinity-8B &
0.0  & 0.001 & 126.47 & 0.271 &
0.0  & 0.0 & 149.95 & 0.241 \\
% 8B-ours
\lightredrow\textbf{+Ours} &
0.0 & 0.005 & 109.73 & 0.270 &
0.0 & 0.013 & 139.73 & 0.235 \\
\bottomrule
\end{tabular}
}
\vspace{-3mm}
\caption{Quantitative comparison between the vanilla model and our \textbf{DiverseVAR} under multiple sampling runs on AFHQ and CelebA-HQ.}
\vspace{-2mm}
\label{tab:fid_AFHQ}%\vspace{-2mm}
\end{table}

% afhq figure

\begin{table*}[tb]\small
\renewcommand{\arraystretch}{0.8} % 控制行距（行高）
\tabcolsep=0.1cm % 控制列间距
\centering
% \resizebox{\linewidth}{!}{
\begin{tabular}{c|c|c|cccc|ccc}
\toprule
{Dataset} & {\multirow{4}{*}{\makecell{Steps$\downarrow$}}} & {\multirow{4}{*}{\makecell{Params$\downarrow$}}} 
& \multicolumn{4}{c|}{{GenEval}} 
& \multicolumn{3}{c}{{DPG}} \\
\cmidrule{1-1}\cmidrule{4-10}
\diagbox{{Method}}{{Metrics}} & & & 
{{Two Obj.}} & {{Position}} & {{Color Attri.}} & {{Overall~$\uparrow$}} &
{{Global}} & {{Relation}} & {{Overall~$\uparrow$}} \\
\midrule
% \midrule
SDXL~\cite{podell2023sdxl} & 40 & 2.6B & 0.74 & 0.15 & 0.23 & 0.55 & 83.27 & 86.76 & 74.65 \\
LlamaGen~\cite{sun2024llamagen} & 1024 & 0.8B & 0.34 & 0.07 & 0.04 & 0.32 & -- & -- & 65.16 \\
Show-o~\cite{xie2024showo} & 1024 &1.3B & 0.80 & 0.31 & 0.50 & 0.68 & -- & -- & 67.48 \\
PixArt-Sigma~\cite{chen2024pixartSigma} & 20 & 0.6B & 0.62 & 0.14 & 0.27 & 0.55 & 86.89 & 86.59 & 80.54 \\
SD3-medium~\cite{esser2024scaling} & 28 & 2.0B & 0.74 & 0.34 & 0.36 & 0.62 & - & - & - \\
DALL-E 3~\cite{DALLE3} & - & -- & -- & -- & -- & 0.67 & 90.97 & 90.58 & 83.50 \\
Emu3~\cite{wang2024emu3} & & 8.5B & 0.81 & 0.49 & 0.45 & 0.66 & -- & -- & 81.60 \\

% \midrule
HART~\cite{tang2024hart} & 14 & 0.7B & 0.62 & 0.13 & 0.18 & 0.51 & -- & -- & 80.89 \\
% \grayrow \textbf{+ Ours (stage1)} & 0.7B & -- & -- & -- & -- & -- & -- & -- \\
% \grayrow \textbf{+ Ours (stage2)} & 0.7B & -- & -- & -- & -- & -- & -- & -- \\
\midrule

% 2B
Infinity-2B~\cite{Infinity} & 13 & 2.0B & 0.84 & 0.45 & 0.55 & 0.73 & 84.80 & 93.04 & 82.97 \\
% stage1
% \grayrow \textbf{+ Ours (stage1)} & 2.0B & 0.81 & 0.39 & 0.53 & 0.68 & 83.89 & 92.45 & 83.10 \\
% stage2
\lightredrow \textbf{+ Ours } & 13 & 2.0B & 0.85 & 0.41 & 0.53 & 0.70 & 85.11 & 92.26 & 83.02 \\

% 8B
Infinity-8B~\cite{Infinity} & 13 & 8.0B & 0.90 & 0.61 & 0.68 & 0.79 & 85.10 & 94.50 & 86.60 \\
% 8B_ours
\lightredrow \textbf{+ Ours } & 13 & 8.0B & 0.89 & 0.59 & 0.66 & 0.76 & 84.80 & 94.93 & 86.78 \\

\bottomrule
\end{tabular}
% }
\vspace{-3mm}
\caption{Quantitative comparisons of perceptual quality on GenEval and DPG benchmarks.}
\vspace{-4mm}
\label{tab:bench}
\end{table*}

\minisection{Overall Comparison.}
We evaluate the performance of our method against the vanilla VAR model~\cite{Infinity}, diffusion-based models~\cite{DALLE3,podell2023sdxl,chen2024pixartSigma}, and AR models~\cite{xie2024showo,wang2024emu3,tang2024hart,sun2024llamagen}.
\cref{tab:bench} summarizes the comparison results on the GenEval~\citep{ghosh2023geneval} and DPG~\citep{DPG-bench} benchmarks. As shown in~\cref{tab:bench}, our method surpasses most competing approaches except, except DALL-E 3 on the DPG benchmark, while maintaining performance on par with the vanilla VAR model.
Specifically, both our method and Infinity achieve over 0.7 on GenEval and around 83.0 on DPG, outperforming most competing methods except DALL-E 3, which demonstrates strong overall performance and consistency with the vanilla Infinity.
% model.

These results indicate that our method generates images with significantly higher diversity while maintaining text-image alignment and visual fidelity compared to the vanilla Infinity model.
See \textit{\textcolor{blue}{Suppl.}} for additional results.

%% ablation table gen_eval (modified, without Params)
\begin{table}[tb]
\renewcommand{\arraystretch}{1} % 控制行距（行高）
\tabcolsep=0.01cm % 控制列间距
\centering
\resizebox{\linewidth}{!}{
\begin{tabular}{c|cccc|ccc}
\toprule
{Dataset} 
& \multicolumn{4}{c|}{{GenEval}} 
& \multicolumn{3}{c}{{DPG}} \\
\cmidrule{1-1}\cmidrule{2-8}
Method &
{{Two Obj.}} & {{Pos.}} & {{Color Attri.}} & {{OA$\uparrow$}} &
{{Global}} & {{Relation}} & {{OA$\uparrow$}} \\
\midrule
% ours-stage1 (抑制->VAR) 
\circlednum{1} SSR$^\dag$ & 0.81 & 0.39 & 0.53 & 0.68 
& 83.89 & 92.45 & 83.01 \\

% 增强->VAR->抑制 (ablation2)
\circlednum{2} SAR$^\dag$+SSR$^\ddag$ & 0.84 & 0.40 & 0.53 & 0.69 
& 84.19 & 92.57 & 82.79 \\

% 抑制->增强->VAR (ablation3)
\circlednum{3} SSR$^\dag$+SAR$^\dag$ & 0.83 & 0.41 & 0.51 & 0.68 
& 82.67 & 92.80 & 82.77 \\

% 增强->抑制->VAR (ablation4)
\circlednum{4} SAR$^\dag$+SSR$^\dag$  & 0.81 & 0.41 & 0.51 & 0.68 
& 82.97 & 91.37 & 82.72 \\

% ours-stage2 (抑制->VAR->增强)
\lightredrow  \makecell{\circlednum{5} SSR$^\dag$+SAR$^\ddag$\\(\textbf{Ours})} & \textbf{0.85} & \textbf{0.41} & \textbf{0.53} & \textbf{0.70 }
& \textbf{85.11} & 92.26 & \textbf{83.02} \\
\bottomrule
\end{tabular}
}
\vspace{-3mm}
\caption{Ablation study of each component in \textbf{DiverseVAR}, evaluating perceptual quality on the GenEval and DPG benchmarks. Components include \textbf{SSR} and \textbf{SAR}. 
% $^\dag$ indicates that the operation is applied to the input feature $\widetilde{\boldsymbol{F}}_{k-1}$, while $^\ddag$ indicates that the operation is applied to the output feature $\boldsymbol{F}_k^o$.
}
\vspace{-2mm}
\label{tab:ablation_bench_no_params}
\end{table}

\begin{table}[tb]
\renewcommand{\arraystretch}{1} % 行高
\tabcolsep=0.03cm % 列宽
\centering
\resizebox{1\linewidth}{!}{
\begin{tabular}{c|ccc|c|ccc|c}
\toprule
{Dataset}
& \multicolumn{4}{c|}{{COCO2014-30K}} & \multicolumn{4}{c}{{COCO2017-5K}} \\
% \cmidrule{1-1}\cmidrule{5-18}
\midrule
Method & {{Recall$\uparrow$}} & {{Cov.$\uparrow$}} &
{{FID$\downarrow$}} & {{CLIP$\uparrow$}} &
{{Recall$\uparrow$}}  & {{Cov.$\uparrow$}} & 
{{FID$\downarrow$}} & {{CLIP$\uparrow$}}  \\
\midrule
\circlednum{1} SSR$^\dag$ &
 0.375 & 0.689 & 23.77 & 0.313 &
 0.440 & 0.858 & 34.65 & 0.313 \\
\circlednum{2} SAR$^\dag$+SSR$^\ddag$ &
 0.352 & 0.686 & 24.98 & 0.313 &
0.456  & 0.857 & 35.36 & 0.313 \\
\circlednum{3} SSR$^\dag$+SAR$^\dag$ &
 0.380 & 0.694 & 23.18 & 0.313 &
0.461  & 0.861 & 33.88 & 0.313  \\
\circlednum{4} SAR$^\dag$+SSR$^\dag$ & 
 0.374 & 0.688 & 23.14 & 0.313 &
 0.471 & 0.871 & 33.85 & 0.313 \\
\lightredrow \makecell{\circlednum{5} SSR$^\dag$+SAR$^\ddag$\\(\textbf{Ours})}&
\textbf{0.385}  & \textbf{0.690} & \textbf{22.96} & 0.313 & 
\textbf{0.480} & {0.860} & \textbf{33.39} & 0.313 \\
\bottomrule
\end{tabular}
}
\vspace{-3mm}
\caption{Ablation study of each component in \textbf{DiverseVAR}, evaluating perceptual quality on the COCO2014 (30K prompts) and COCO2017 (5K prompts) benchmarks. Components include \textbf{SSR} and \textbf{SAR}. 
% $^\dag$ indicates that the operation is applied to the input feature $\widetilde{\boldsymbol{F}}_{k-1}$, while $^\ddag$ indicates that the operation is applied to the output feature $\boldsymbol{F}_k^o$.
}
\vspace{-2mm}
\label{tab:ablation_fid_clip}\vspace{-2mm}
\end{table}

\subsection{Additional Analysis}
\minisection{Ablation Study.}
Based on our analysis and observation, \textbf{DiverseVAR} applies \textit{Soft-Suppression Regularization} (i.e., \textbf{SSR}) to the input feature $\widetilde{\boldsymbol{F}}_{k-1}$, and \textit{Soft-Augmentation Regularization} (i.e., \textbf{SAR}) to the output feature $\boldsymbol{F}_k^o$ (i.e., \circlednum{5} in both~\cref{tab:ablation_bench_no_params,tab:ablation_fid_clip}).
We perform an ablation study on these components, exploring alternative designs and comparing performance. 
The ablated designs include: \circlednum{1} SSR$^\dag$: The SAR is removed, and only the SSR is applied to the input feature $\widetilde{\boldsymbol{F}}_{k-1}$. \circlednum{2} SAR$^\dag$+SSR$^\ddag$: Apply the SAR to the input feature $\widetilde{\boldsymbol{F}}_{k-1}$, and the SSR to the output feature $\boldsymbol{F}_k^o$. \circlednum{3} SSR$^\dag$+SAR$^\dag$: First, apply SSR and then SAR to the input feature $\widetilde{\boldsymbol{F}}_{k-1}$. \circlednum{4} SAR$^\dag$+SSR$^\dag$: First, apply SAR and then SSR to the input feature $\widetilde{\boldsymbol{F}}_{k-1}$. 
Here, $^\dag$ indicates that the operation is applied to the input feature $\widetilde{\boldsymbol{F}}_{k-1}$, while $^\ddag$ indicates that the operation is applied to the output feature $\boldsymbol{F}_k^o$.
Performance results and comparisons are presented in~\cref{tab:ablation_bench_no_params,tab:ablation_fid_clip}.
We observe that our method (i.e., \circlednum{5}) outperforms other ablated designs (i.e., \circlednum{1}-\circlednum{4}) in both perceptual quality and diversity, except for the Relation on DPG and the Coverage metric on COCO2017.
See \textit{\textcolor{blue}{Suppl.}} for additional ablation.

\begin{figure}
    \centering
    \includegraphics[width=0.92\linewidth]{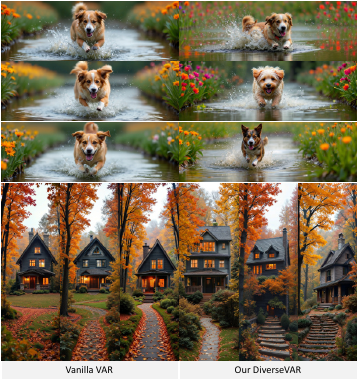}\vspace{-2mm}
    \caption{Qualitative comparison between the vanilla model and our method. Our \textbf{DiverseVAR} facilitates generation with diverse aspect ratios. See \textit{\textcolor{blue}{Suppl.}} for the text prompts used.}\vspace{-6mm}
    \label{fig:varying_aspect_ratios}
\end{figure}

\minisection{Additional Results.}
%% fig: varying_aspect_ratios
The Infinity~\citep{Infinity} model originally supports image generation with varying aspect ratios, and our method, \textbf{DiverseVAR}, preserves this property. As shown in~\cref{fig:varying_aspect_ratios}, when combined with \textbf{DiverseVAR}, it continues to facilitate efficient image generation, demonstrating that our proposed method can be easily extended to generate images with diverse aspect ratios.

%% prompts
% A dog running through water surrounded by flowers.
% A cozy house surrounded by autumn trees.

% \minisection{Creative Generation.}

% \begin{figure}
%     \centering
%     \includegraphics[width=1\linewidth]{figure/figure10_creative_study_compressed.pdf}\vspace{-2mm}
%     \caption{Our method can also be applied to creative generation.}\vspace{-4mm}
%     \label{fig:Creative Generation.}
% \end{figure}
%% prompts
% a creative log cabin in the forest.
% A creative library beside an ancient tree filled with wisdom.
% A creative small cottage surrounded by blooming flowers.
% A creative temple beside an ancient tree with timeless charm.
% A creative tower standing gracefully beside an ancient tree.

\section{Conclusion}

In this work, we explore the factors that influence diversity in text-to-image VAR models and find that the pivotal component plays a critical role in shaping diversity formation at early scales. 
Building on this insight, we introduce \textbf{DiverseVAR}, a simple yet effective framework that unleashes the intrinsic generative diversity of VAR models while preserving image fidelity, through input-level suppression and output-level augmentation of the pivotal component. We conduct extensive experiments and demonstrate that our approach effectively improves generative diversity while preserving image quality and text–image alignment.

% \newpage

{
    \small
    \bibliographystyle{unsrt}
    \bibliography{longstrings,main}
}

\clearpage
\setcounter{page}{1}
\maketitlesupplementary

\renewcommand{\thesection}{\Alph{section}} % 将节号改为大写字母
\setcounter{section}{0}

\setcounter{table}{0}
\setcounter{figure}{0}
\renewcommand{\thefigure}{S\arabic{figure}}
\renewcommand{\thetable}{S\arabic{table}}

\section{Overview}
This supplementary material provides additional implementation details, algorithmic descriptions, and extended experiments to support the main paper. Specifically, it includes:
\begin{itemize} 
    \item \textbf{Implementation Details (\cref{sec:imdetails}}): Detailed configuration of datasets, metrics, and baseline implementations used in our experiments.
    \item \textbf{Algorithm Detail of \ourmethod (\cref{sec:algorithm}}): Complete pseudocode and procedural explanation of our proposed method for reproducibility.
    \item \textbf{Additional Analysis (\cref{sec:add_ana}}): Ablation studies on different design choices, in-depth analysis of scale and block configurations, and additional qualitative comparisons demonstrating image generation quality and diversity. 
\end{itemize}

\section{Implementation Details}
\label{sec:imdetails}

\subsection{Configure}

\minisection{Datasets.}
We utilize the AFHQ~\cite{choi2020starganv2} and CelebA-HQ~\cite{karras2017progressive} datasets to evaluate diversity under multiple sampling scenarios, as they contain diverse animal and human face images, respectively. We employ text prompts in the following formats: “A face of a $<$cat/dog/wild animal$>$” and “A face of a $<$man/woman$>$”.

\minisection{Metrics.}
We leverage the code from the popular GitHub repository “StudioGAN”~\citep{kang2023studiogan, Kang2021RebootingAA, Kang2020ContraGANCL}\footnote{\url{https://github.com/POSTECH-CVLab/PyTorch-StudioGAN}}
 to compute two metrics: Recall~\citep{kynkaanniemi2019improved} and Coverage~\citep{naeem2020reliable}.
For FID~\citep{heusel2017gans} and CLIP Score~\citep{hessel2021clipscore}, we employ the official evaluation code from GigaGAN~\citep{kang2023scaling}\footnote{\url{https://github.com/lucidrains/gigagan-pytorch}}
We use the official evaluation code for GenEval~\citep{ghosh2023geneval}\footnote{\url{https://github.com/djghosh13/geneval}}
 (also provided in the official Infinity code) and DPG~\citep{DPG-bench}\footnote{\url{https://github.com/TencentQQGYLab/ELLA}}

\minisection{Baseline Implementations.}
We use the official implementation of Infinity-2B and Infinity-8B~\cite{Infinity}\footnote{\url{https://github.com/FoundationVision/Infinity}}. For Infinity-2B, we use all the hyperparameters at their default settings. For Infinity-8B, due to computational resource limitations, using A100 40GB GPUs leads to out-of-memory (OOM) errors. Therefore, we conduct both qualitative and quantitative evaluations on the generation results at scale 48.

\subsection{The Pivotal and Auxiliary Token}
\label{subsec:patoken}
In our \textbf{\textit{Observation 2}}, we employ the defined pivotal score and the Maximum Distance to Chord (MDC) to identify the \textit{pivotal} and \textit{auxiliary} token.
Specifically, we follow~\cite{guo2025fastvar} and define the pivotal score $s_{k,i} = \|\widetilde{\boldsymbol{F}}_{k-1,i}-\bar{\boldsymbol{F}}_{k-1}\|_2$ using the L2 norm,
% \begin{equation} 
% s_k = \|\boldsymbol{F}_k-\bar{\boldsymbol{F}}_k\|_2,
% \label{eq:sk}
% \end{equation}
where $\bar{\boldsymbol{F}}_{k-1}$ represents the mean feature map obtained by averaging $\widetilde{\boldsymbol{F}}_{k-1}$ across scale dimensions, and $\widetilde{\boldsymbol{F}}_{k-1,i}$ denotes each token along the scale dimension.
% Based on their pivotal scores, all tokens in the feature map are ranked and then categorized into \textit{pivotal} and \textit{auxiliary} tokens, with the partition determined by the elbow point computed using the Maximum Distance to Chord (MDC)~\cite{douglas1973algorithms}
Then, to determine the boundary between \textit{pivotal} and \textit{auxiliary} tokens, we follow the MDC method~\cite{douglas1973algorithms}.
Given the sorted sequence $\{s_{k,i_n}\}_{n=1}^{L}$ in descending order, we construct a chord connecting the endpoints $(1, s_{k,i_1})$ and $(L, s_{k,i_L})$, where $L=S_k$ is the scale dimension.
The perpendicular distance of each intermediate point $(i_n, s_{k,i_n})$ to this chord is computed as
% \begin{equation}
% d_n = 
% \frac{\big| (s_{k,i_L - s_{k,i_1}) \cdot i_n 
% - (L - 1) \cdot s_{k,i_n} 
% + L \cdot s_{k,i_1} - s_{k,i_L \big|}
% {\sqrt{(s_{k,i_L - s_{k,i_1})^2 + (i_L - 1)^2}}.
% \label{eq:mdc}
% \end{equation}
% The index $n^* = \arg\max_n d_n$ corresponding to the maximum distance is defined as the \textbf{elbow point}, which separates the tokens into
% \begin{equation}
% \mathcal{T}_{\text{pivotal}} = \{1, \ldots, i_{n^*}\}, \quad
% \mathcal{T}_{\text{auxiliary}} = \{i_{n^*}+1, \ldots, L\}.
% \end{equation}
% \begin{equation}
% d_{i_n} = 
% \frac{\big| (s_{k,i_L} - s_{k,i_1}) \cdot i_n 
% - (L - 1) \cdot s_{k,i_n} 
% + L \cdot s_{k,i_1} - s_{k,i_L} \big|}
% {\sqrt{(s_{k,i_L} - s_{k,i_1})^2 + (L - 1)^2}}.
% \label{eq:mdc}
% \end{equation}
{
\small
\begin{equation}
d_{i_n} = 
\frac{\big| (s_{k,i_L} - s_{k,i_1}) \cdot i_n 
- (L - 1) \cdot s_{k,i_n} 
+ L \cdot s_{k,i_1} - s_{k,i_L} \big|}
{\sqrt{(s_{k,i_L} - s_{k,i_1})^2 + (L - 1)^2}}.
\label{eq:mdc}
\end{equation}
}

The index $i_n^* = \arg\max_{i_n} d_{i_n}$ corresponding to the maximum distance is defined as the {elbow point}, which separates the tokens into
\begin{equation}
\mathcal{T}_{\text{pivotal}} = \{1, \ldots, i_n^*\}, \quad
\mathcal{T}_{\text{auxiliary}} = \{i_n^*+1, \ldots, L\}.
\end{equation}
Tokens before the elbow point exhibit higher deviation from the mean feature and are thus considered \textit{pivotal}, while those after are treated as \textit{auxiliary}.
% contributing less to semantic representation but useful for contextual smoothing.

\subsection{Text Prompts}
% Below are the text prompts used for image generation in the paper.
We list the text prompts used for image generation in this paper below.

\minisection{\cref{fig:teaser}}: ``A man in a clown mask eating a donut'', ``A cat wearing a Halloween costume'', ``Golden Gate Bridge at sunset, glowing sky, dramatic perspective, vivid colors'', ``A palace under the sunset'', ``A cool astronaut floating in space'', and ``A cat riding a skateboard down a hill''.
% A man in a clown mask eating a donut.
% A cat wearing a Halloween costume.
% Golden Gate Bridge at sunset, glowing sky, dramatic perspective, vivid colors.
% A palace under the sunset.
% A cool astronaut floating in space.
% A cat riding a skateboard down a hill.

\minisection{\cref{fig:Observation_1}}: ``a green train is coming down the tracks''.
% prompt: a green train is coming down the tracks

\minisection{\cref{fig:Observation_2}}: ``A bird perched on a tree branch in the rain''.
% prompt: a bird perched on a tree branch in the rain

\minisection{\cref{fig:zeroing_s0}}: ``A bear fishing with a stick by a calm river''.
% % prompt: A bear fishing with a stick by a calm river.

\minisection{\cref{fig:step2}}: ``A hot air balloon floating above the clouds''.
% % prompt: A hot air balloon floating above the clouds.

\minisection{\cref{fig:afhq_cat}}: ``The face of a cat''.
% Prompt: The face of a cat.

\minisection{\cref{fig:varying_aspect_ratios}}: ``A dog running through water surrounded by flowers'' and ``A cozy house surrounded by autumn trees''.
%% prompts
% A dog running through water surrounded by flowers.
% A cozy house surrounded by autumn trees.

\minisection{\cref{fig:sup_visualizaiton_a}}: ``A very cute cat near a bunch of birds'', ``A cat standing on a hill'', ``A photo of a cute rabbit holding a cup of coffee in a café'', ``A cinematic shot of a little pig priest wearing sunglasses'', ``A dog covered in vines'', and ``Cute grey cat, digital oil painting by Monet''.
%% prompt 1-6
% A very cute cat near a bunch of birds.
% A cat standing on a hill.
% A photo of a cute rabbit holding a cup of coffee in a café.
% A cinematic shot of a little pig priest wearing sunglasses.
% A dog covered in vines.
% Cute grey cat, digital oil painting by Monet.

\minisection{\cref{fig:sup_visualizaiton_b}}: ``Editorial photoshoot of an old woman, high fashion 2000s fashion'', ``An astronaut riding a horse on the moon, oil painting by Van Gogh'', ``Full body shot, a French woman, photography, French streets'', ``A boy and a girl fall in love'', ``An abstract portrait of a pensive face, rendered in cool shades of blues, purples, and grays'', and ``Cute boy, hair looking up to the stars, snow, beautiful lighting, painting style by Abe Toshiyuki''.
%% prompt 7-12
% Editorial photoshoot of an old woman, high fashion 2000s fashion.
% An astronaut riding a horse on the moon, oil painting by Van Gogh.
% Full body shot, a French woman, photography, French streets.
% A boy and a girl fall in love.
% An abstract portrait of a pensive face, rendered in cool shades of blues, purples, and grays.
% Cute boy, hair looking up to the stars, snow, beautiful lighting, painting style by Abe Toshiyuki.

\minisection{\cref{fig:sup_visualizaiton_c}}: ``A table with a light on over it'', ``A library filled with warm yellow light'', ``A villa standing on a hill'', ``A train crossing a bridge over a canyon'', ``A bridge stretching over a calm river'', and ``A temple surrounded by flowers''.
% A table with a light on over it.
% A library filled with warm yellow light.
% A villa standing on a hill.
% A train crossing a bridge over a canyon.
% A bridge stretching over a calm river.
% A temple surrounded by flowers.

\section{Algorithm detail of \ourmethod}\label{sec:algorithm}

\begin{algorithm}[H]
\small %\normalsize % \footnotesize
\renewcommand{\arraystretch}{0.1}
\tabcolsep=0.01cm
\SetAlgoLined
\SetKwInOut{KwInput}{Input}
\SetKwInOut{KwOutput}{Output}
\KwInput{
Scales $\{S_1,S_2,\cdots,S_K\}$. Scales for $\ourmethod$ consists of $m$ scales $\{l_1,\cdots,l_m\}$. Scales for the vanilla process $\{S_1,S_2,\cdots,S_K\}\setminus\{l_1,\cdots,l_m\}$.
The VAR model $\mathcal{\phi}$, the image decoder $\mathcal{D}$, and the quantizer $\mathcal{Q}$. %The quantizer $\mathcal{Q}$, which typically includes a codebook $Z\in\mathbb{R}^{V\times d}$ containing $V$ vectors.
}
\KwOutput{The final diverse output $\mathbf{I}$.}
\vspace{1mm} \hrule \vspace{1mm}
\noindent $\boldsymbol{F}_0=0$\;\gray{;} 

\noindent $\widetilde{\boldsymbol{F}}_0 = \langle \text{SOS} \rangle \in \mathbb{R}^{1 \times 1 \times d}$ \gray{\tcp*{The start token~\citep{Infinity}}} 

% \noindent \gray{\tcp{\textit{DiverseVAR}}}
\For{$k=1,\cdots,K$}{
    \If{$S_k \in\{S_1,S_2,\cdots,S_K\}\setminus\{l_1,\cdots,l_m\}$}{
        $\boldsymbol{F}_k^o = \phi(\widetilde{\boldsymbol{F}}_{k-1})$ \gray{\tcp*{\cref{fig:framework} (Left)}} 
    }
    \Else{ 
        % \gray{\tcp{\small \textit{Soft-Suppression Regularization}}}
        $\hat{\boldsymbol{F}}_{k-1}\gets\widetilde{\boldsymbol{F}}_{k-1}$ \gray{\tcp*{SSR: \cref{eq:ssr}}}
        
        $\hat{\boldsymbol{F}}_k^o = \phi(\hat{\boldsymbol{F}}_{k-1})$ \gray{\tcp*{\cref{fig:framework} (Right)}}

        % \gray{\tcp{\small \textit{Soft-Augmentation Regularization}}}
        ${\boldsymbol{F}}_k^o\gets\hat{\boldsymbol{F}}_k^o$ \gray{\tcp*{SAR: \cref{eq:sar}}}
    }
    $\boldsymbol{R}_k=\mathcal{Q}{(\boldsymbol{F}_k^o)}$\;\gray{;}
    
    $\boldsymbol{F}_k = \boldsymbol{F}_{k-1}  + \mathrm{up}(\boldsymbol{R}_k,(h_K,w_K))$ \gray{\tcp*{\cref{eq:var_upf}}}

    $\widetilde{\boldsymbol{F}}_{k}=\mathrm{down}(\boldsymbol{F}_k,(h_k,w_k))$ \gray{\tcp*{\cref{eq:var_downf}}} 
}

$\mathbf{I}=\mathcal{D}(\boldsymbol{F}_K)$

\textbf{Return} The final generated image $\mathbf{I}$
\caption{: \normalsize \ourmethod}
\label{alg:diversevar}
\end{algorithm}

%% 补充实验部分（1）
\begin{table*}[t]\footnotesize %\small
\renewcommand{\arraystretch}{0.8} % 控制行距（行高）
\tabcolsep=0.03cm % 控制列间距
\centering
\begin{tabular}{c|cccc|ccc|cccc|cccc}
\toprule
{Dataset} 
& \multicolumn{4}{c|}{{GenEval}} 
& \multicolumn{3}{c|}{{DPG}} 
& \multicolumn{4}{c|}{{COCO2014-30K}} 
& \multicolumn{4}{c}{{COCO2017-5K}} \\
\cmidrule{1-1}\cmidrule{2-16}
\diagbox{Scales}{{Metrics}} &
{{Two Obj.}} & {{Position}} & {{Color Attri.}} & {{Overall~$\uparrow$}} &
{{Global}} & {{Relation}} & {{Overall~$\uparrow$}} &
{{Recall$\uparrow$}} & {{Cov.$\uparrow$}} & {{FID$\downarrow$}} & {{CLIP$\uparrow$}} &
{{Recall$\uparrow$}} & {{Cov.$\uparrow$}} & {{FID$\downarrow$}} & {{CLIP$\uparrow$}} \\
\midrule

$l_i=\varnothing$ (Vanilla) & 0.84 & 0.45 & 0.55 & 0.73 & 84.80 & 93.04 & 82.97 & 0.316  & 0.651 & 28.48 & 0.313 & 0.408  & 0.832 & 39.01 & 0.313 \\
\midrule
$l_i\in\{2\}$ & 0.85 & 0.48 & 0.52 & 0.733 & 83.28 & 92.06 & 83.018 & 0.321 & 0.674 & 27.26 & 0.313 & 0.417 & 0.843 & 37.80 & 0.313 \\
$l_i\in\{4\}$ & 0.83 & 0.44 & 0.58 & 0.716 & 83.28 & 92.49 & 83.256 & 0.333 & 0.660 & 27.11 & 0.314 & 0.430 & 0.833 & 37.64 & 0.313 \\
$l_i\in\{6\}$ & 0.85 & 0.42 & 0.52 & 0.711 & 85.41 & 92.72 & 83.254 & 0.337 & 0.667 & 26.64 & 0.314 & 0.415 & 0.844 & 36.87 & 0.313 \\
$l_i\in\{8\}$ & 0.84 & 0.45 & 0.52 & 0.725 & 83.89 & 92.49 & 82.891 & 0.329 & 0.662 & 27.35 & 0.314 & 0.422 & 0.838 & 37.92 & 0.313 \\
$l_i\in\{2,4\}$ & 0.83 & 0.44 & 0.52 & 0.712 & 82.06 & 92.22 & 82.859 & 0.340 & 0.680 & 25.43 & 0.313 & 0.435 & 0.851 & 35.92 & 0.312 \\
\lightredrow \textbf{$l_i\in\{4,6\}$ (Ours)} & 0.85 & 0.41 & 0.53 & 0.70 & 85.11 & 92.26 & 83.02 & 0.385  & 0.690 & 22.96 & 0.313 & 
0.480 & 0.860 & 33.39 & 0.313 \\
$l_i\in\{6,8\}$ & 0.80 & 0.40 & 0.47 & 0.677 & 83.89 & 92.45 & 82.996 & 0.366 & 0.693 & 23.24 & 0.313 & 0.452 & 0.855 & 33.84 & 0.313 \\
$l_i\in\{2,4,6\}$ & 0.77 & 0.40 & 0.48 & 0.669 & 83.58 & 92.26 & 82.388 & 0.432 & 0.710 & 20.05 & 0.312 & 0.513 & 0.876 & 30.79 & 0.311 \\
$l_i\in\{4,6,8\}$ & 0.75 & 0.34 & 0.42 & 0.622 & 83.58 & 92.37 & 81.630 & 0.499 & 0.700 & 15.85 & 0.310 & 0.567 & 0.878 & 26.35 & 0.310 \\
$l_i\in\{2,4,6,8\}$ & 0.70 & 0.35 & 0.35 & 0.593 & 82.37 & 92.45 & 80.749 & 0.544 & 0.687 & 14.70 & 0.306 & 0.611 & 0.857 & 25.67 & 0.304 \\
\bottomrule
\end{tabular}
\vspace{-3mm}
\caption{Ablation study of different scales $l_i$ on image generation quality (GenEval and DPG) and diversity (COCO 2014 and COCO 2017).}
\vspace{-2mm}
\label{tab:ablation_study_1}
\end{table*}

%% 补充实验部分（2）
\begin{table*}[t]\footnotesize %\mysmallerfont %\footnotesize %\small
\renewcommand{\arraystretch}{0.8} % 控制行距（行高）
\tabcolsep=0.03cm % 控制列间距
\centering
\begin{tabular}{c|cccc|ccc|cccc|cccc}
\toprule
{Dataset} 
& \multicolumn{4}{c|}{{GenEval}} 
& \multicolumn{3}{c|}{{DPG}} 
& \multicolumn{4}{c|}{COCO2014-30K} 
& \multicolumn{4}{c}{{COCO2017-5K}} \\
\cmidrule{1-1}\cmidrule{2-16}
\diagbox{Blocks}{{Metrics}} &
{{Two Obj.}} & {{Position}} & {{Color Attri.}} & {{Overall~$\uparrow$}} &
{{Global}} & {{Relation}} & {{Overall~$\uparrow$}} &
{{Recall$\uparrow$}} & {{Cov.$\uparrow$}} & {{FID$\downarrow$}} & {{CLIP$\uparrow$}} &
{{Recall$\uparrow$}} & {{Cov.$\uparrow$}} & {{FID$\downarrow$}} & {{CLIP$\uparrow$}} \\
\midrule
$i=\varnothing$ & 0.84 & 0.45 & 0.55 & 0.73 & 84.80 & 93.04 & 82.97 & 0.316  & 0.651 & 28.48 & 0.313 & 0.408  & 0.832 & 39.01 & 0.313 \\
\midrule
\lightredrow \makecell{$i\in\{0,1,2,3,4,5,6,7\}$\\\textbf{(Ours)}} & 0.85 & 0.41 & 0.53 & 0.70 & 85.11 & 92.26 & 83.02 & 0.385  & 0.690 & 22.96 & 0.313 & 
0.480 & 0.860 & 33.39 & 0.313 \\
\midrule
$i\in\{1,2,3,4,5,6,7\}$ & 0.80 & 0.39 & 0.50 & 0.678 & 85.41 & 92.57 & 82.93 & 0.370 & 0.691 & 23.77 & 0.314 & 0.456 & 0.867 & 34.38 & 0.313 \\
$i\in\{0,1,2,3,4,5,6\}$ & 0.80 & 0.38 & 0.45 & 0.657 & 83.89 & 92.72 & 82.60 & 0.411 & 0.699 & 21.06 & 0.314 & 0.485 & 0.868 & 31.92 & 0.313 \\
$i\in\{2,3,4,5,6,7\}$ & 0.85 & 0.42 & 0.56 & 0.715 & 83.58 & 92.49 & 82.86 & 0.350 & 0.676 & 25.78 & 0.314 & 0.432 & 0.851 & 36.27 & 0.314 \\
$i\in\{4,5,6,7\}$ & 0.82 & 0.42 & 0.56 & 0.722 & 84.19 & 92.26 & 83.03 & 0.338 & 0.659 & 27.22 & 0.314 & 0.427 & 0.846 & 37.15 & 0.314 \\
$i\in\{1,2,3\}$ & 0.82 & 0.39 & 0.47 & 0.682 & 84.49 & 92.45 & 83.29 & 0.352 & 0.677 & 24.08 & 0.313 & 0.439 & 0.855 & 34.17 & 0.312 \\
\midrule
Model-level & 0.85 & 0.43 & 0.56 & 0.73 & 84.19 & 92.80 & 83.03 & 0.317 & 0.651 & 28.29 & 0.314 & 0.402 & 0.828 & 38.98 & 0.313 \\
\bottomrule
\end{tabular}
\vspace{-3mm}
\caption{Ablation study of different blocks $b_i$ on image generation quality (GenEval and DPG) and diversity (COCO 2014 and COCO 2017).}
\vspace{-2mm}
\label{tab:ablation_study_2}
\end{table*}

%% 补充实验部分（3）:抑制 logits
\begin{table*}[!tp]\footnotesize %\small
\renewcommand{\arraystretch}{0.8} % 控制行距（行高）
\tabcolsep=0.02cm % 控制列间距
\centering
\begin{tabular}{c|cccc|ccc|cccc|cccc}
\toprule
{Dataset} 
& \multicolumn{4}{c|}{{GenEval}} 
& \multicolumn{3}{c|}{{DPG}} 
& \multicolumn{4}{c|}{{COCO2014-30K}} 
& \multicolumn{4}{c}{{COCO2017-5K}} \\
\cmidrule{1-1}\cmidrule{2-16}
\diagbox{{Method}}{{Metrics}} &
{{Two Obj.}} & {{Position}} & {{Color Attri.}} & {{Overall~$\uparrow$}} &
{{Global}} & {{Relation}} & {{Overall~$\uparrow$}} &
{{Recall$\uparrow$}} & {{Cov.$\uparrow$}} & {{FID$\downarrow$}} & {{CLIP$\uparrow$}} &
{{Recall$\uparrow$}} & {{Cov.$\uparrow$}} & {{FID$\downarrow$}} & {{CLIP$\uparrow$}} \\
\midrule
Vanilla & 0.84 & 0.45 & 0.55 & 0.73 & 84.80 & 93.04 & 82.97 & 0.316  & 0.651 & 28.48 & 0.313 & 0.408  & 0.832 & 39.01 & 0.313 \\
\midrule
SSR in \textit{logits} & 0.86 & 0.45 & 0.82 & 0.72 & 83.28 & 91.99 & 83.41 & 0.320 & 0.604 & 28.95 & 0.313 & 0.419 & 0.838 & 38.91 & 0.313 \\
SSR+SAR in \textit{logits} & 0.86 & 0.45 & 0.54 & 0.73 & 83.89 & 92.99 & 83.12 & 0.331 & 0.580 & 29.06 & 0.313 & 0.402 & 0.828 & 39.22 & 0.313 \\
\lightredrow \makecell{SSR+SAR in blocks\\\textbf{(Ours)}} & 0.85 & 0.41 & 0.53 & 0.70 & 85.11 & 92.26 & 83.02 & 0.385  & 0.690 & 22.96 & 0.313 & 
0.480 & 0.860 & 33.39 & 0.313 \\
\bottomrule
\end{tabular}
\vspace{-3mm}
\caption{Ablation study on \textit{logits} for image generation quality (GenEval and DPG) and diversity (COCO 2014 and COCO 2017).}
\vspace{-2mm}
\label{tab:ablation_study_3}
\end{table*}

\section{Additional Analysis}\label{sec:add_ana}

\subsection{Ablation Study of Scales}
Based on our observation and analysis, early scales have a significant impact on the structural formation. Specifically, we apply $\text{our method}$ to the VAR at scales 2 and 4 (i.e., $l_i\in\{4,6\}$).
To quantitatively investigate the impact of scales on image generation quality and diversity, we conducted experiments on GenEval and DPG to compare generation quality, and on COCO 2014 and COCO 2017 to evaluate generation diversity.
As shown in~\cref{tab:ablation_study_1} (1st–4th rows), applying \ourmethod with a single early scale leads to consistent improvements in Recall, Coverage, and FID over the vanilla on COCO 2014 and COCO 2017, demonstrating enhanced diversity.
Meanwhile, the performance on GenEval and DPG shows only marginal degradation, indicating that image generation quality is well preserved.
To further stimulate model diversity, we apply \ourmethod to multiple early scales and their various combinations.
As illustrated in~\cref{tab:ablation_study_1} (5th–10th rows), increasing the number of applied scales leads to higher diversity but causes a rapid degradation in generation quality.
Notably, applying \ourmethod to four early scales (i.e., $l_i\in\{2,4,6,8\}$) yields the highest diversity metrics but the lowest generation quality, with GenEval decreasing by more than 0.1.
Therefore, to achieve a good trade-off between generation quality and diversity, we apply \ourmethod at scales 4 and 6 (i.e., $l_i\in\{4,6\}$), which are used as the default settings.

\subsection{Ablation Study of Blocks}
% Our method operates at the block level (e.g., 8 blocks $b_i$ in the Infinity backbone, where $i\in\{0,1,2,3,4,5,6,7\}$).
Our method operates at the block level, i.e., on 8 blocks $b_i$ within the Infinity backbone, where $i\in \{0, 1, 2, 3, 4, 5, 6, 7\}$. Specifically, \ourmethod is applied across all 8 blocks. To further assess the contribution of different blocks, we perform comparative experiments by selectively applying \ourmethod to specific subsets of blocks. 
As illustrated in~\cref{tab:ablation_study_2} (3rd–7th rows), applying \ourmethod to only partial subsets of blocks results in a significant degradation in either image generation quality or diversity compared to applying it to all blocks.
For example, applying \ourmethod to blocks $b_i$ with $i\in\{0,1,2,3,4,5,6\}$ yields the best diversity performance — achieving superior Recall, Coverage, and FID on COCO 2014 and COCO 2017 — but results in a severe degradation of image quality (\cref{tab:ablation_study_2} (3rd row)). While applying \ourmethod to blocks $b_i$ with $i\in\{4,5,6,7\}$ maintains the generation quality, the resulting diversity metrics remain suboptimal (\cref{tab:ablation_study_2} (6th row)). Therefore, to achieve a good trade-off between generation quality and diversity, we apply \ourmethod at all 8 blocks, which are used as the default settings.

For a more comprehensive comparison, we further apply the method at the model level.
As illustrated in~\cref{tab:ablation_study_2} (last row), this configuration maintains the generation quality (evaluated on GenEval and DPG), with GenEval remaining at 0.73, but yields almost no improvement in diversity (evaluated on COCO 2014 and COCO 2017).

\subsection{Ablation Study on \textit{logits}}
We observe that the \textit{logits} distributions under different samplings are similar in the vanilla VAR, causing diversity collapse.
Our proposed method mitigates this issue by introducing variation into the distributions, thereby encouraging diverse generations while maintaining faithful text–image alignment and high visual quality.
To further examine whether \ourmethod is more effective at the block level than at the logits level, we performed comparative experiments by applying it to the logits (\cref{tab:ablation_study_3} (2nd-3rd rows)).
As shown in~\cref{tab:ablation_study_3}, when \ourmethod is applied to the logits, the generation quality is well preserved — as indicated by a GenEval score of 0.72 and 0.73 on GenEval — while the diversity metrics on COCO 2014 and COCO 2017 show only a slight improvement.

\section{Additional Results}\label{sec:add_res}

\subsection{Diversity Comparison}
\begin{figure*}
    \centering
    \includegraphics[width=\linewidth]{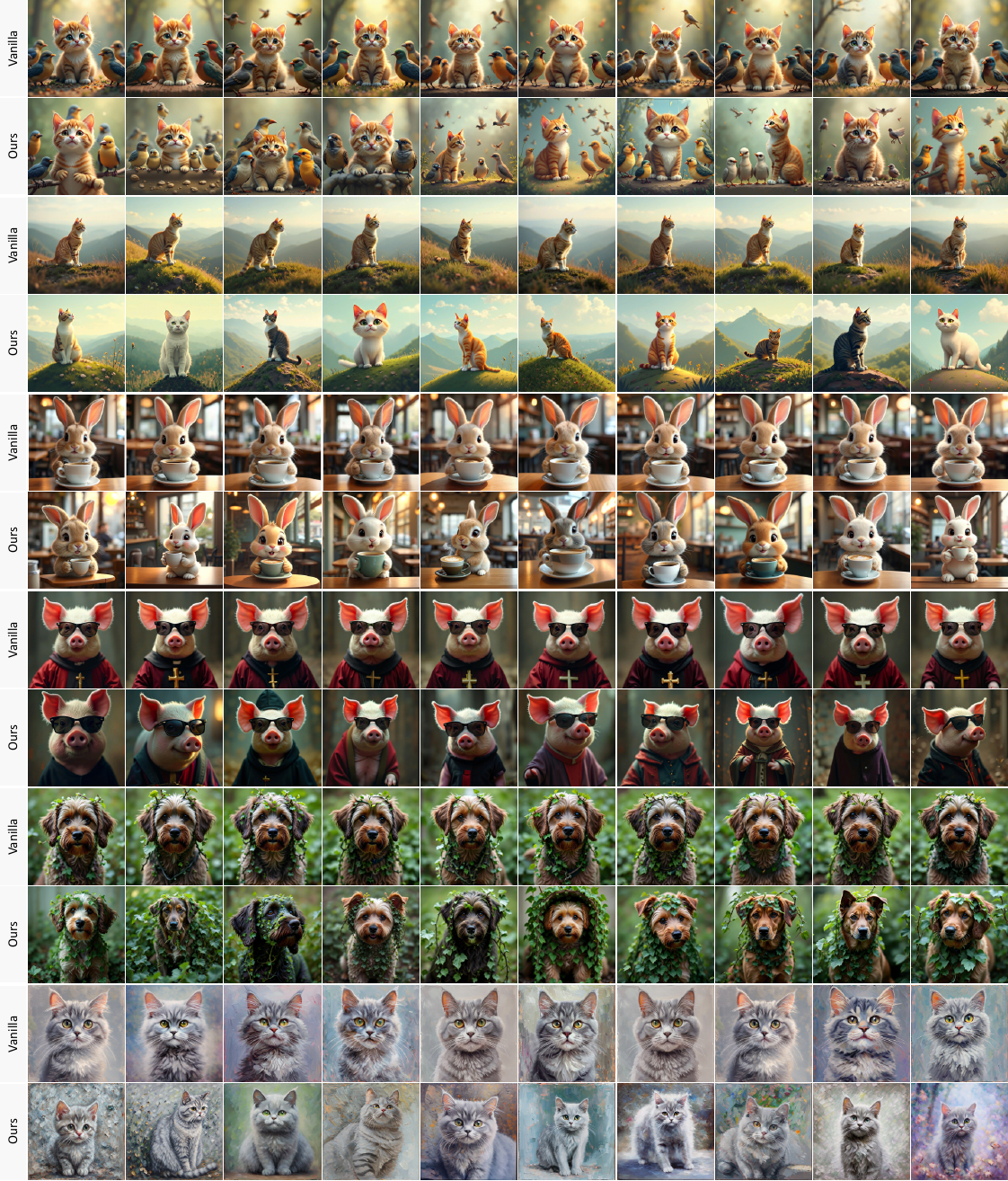}
    \vspace{-1.5\baselineskip}
    \caption{Additional diversity comparison. Our results demonstrate superior diversity.}
    \label{fig:sup_visualizaiton_a}
\end{figure*}

\begin{figure*}
    \centering
    \includegraphics[width=\linewidth]{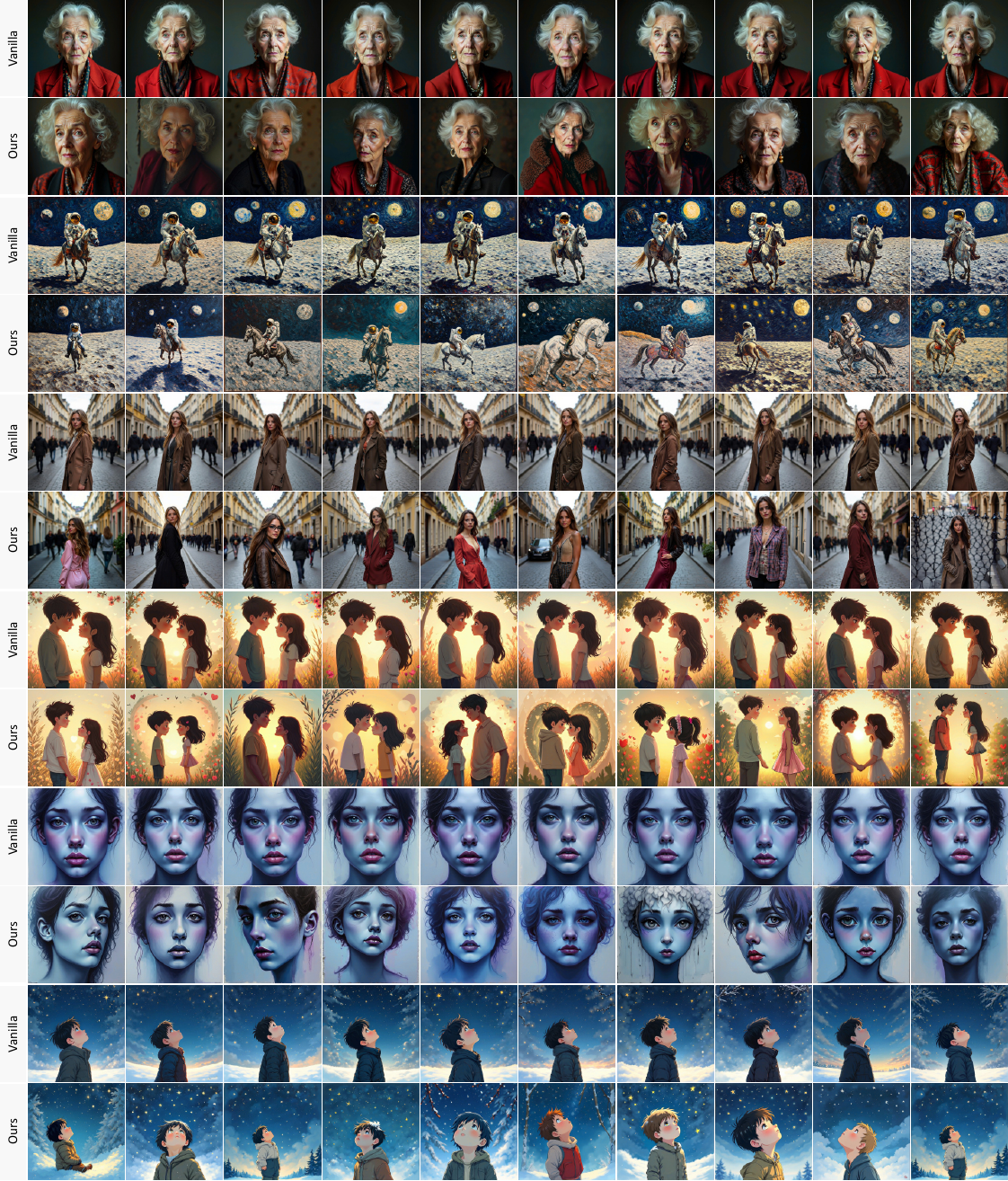}
    \vspace{-1.5\baselineskip}
    \caption{Additional diversity comparison. Our results demonstrate superior diversity.}
    \label{fig:sup_visualizaiton_b}
\end{figure*}

\begin{figure*}
    \centering
    \includegraphics[width=\linewidth]{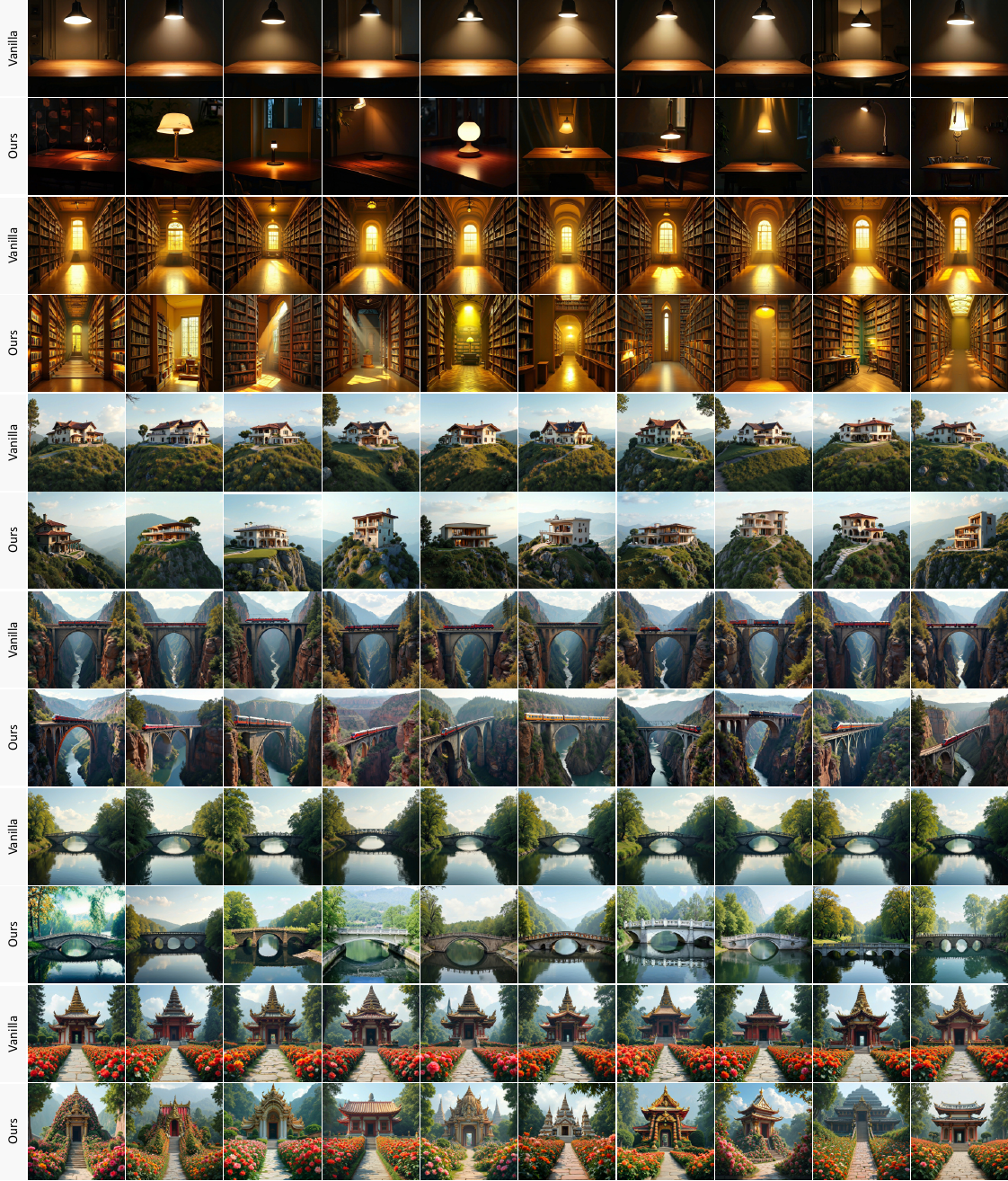}
    \vspace{-1.5\baselineskip}
    \caption{Additional diversity comparison. Our results demonstrate superior diversity.}
    \label{fig:sup_visualizaiton_c}
\end{figure*}

As shown in~\cref{fig:sup_visualizaiton_a,fig:sup_visualizaiton_b,fig:sup_visualizaiton_c}, we present additional qualitative results for diversity comparison.
We observe that the vanilla model tends to produce samples with limited diversity across different text prompts, although it generates high-quality results. In contrast, our method can generate diverse images from multiple samples under different text prompts, while maintaining good text–image alignment, indicating its advantage over the baselines.

%% prompt 1-6
% A very cute cat near a bunch of birds.
% A cat standing on a hill.
% A photo of a cute rabbit holding a cup of coffee in a café.
% A cinematic shot of a little pig priest wearing sunglasses.
% A dog covered in vines.
% Cute grey cat, digital oil painting by Monet.

%% visualization-b
%% prompt 7-12
% Editorial photoshoot of an old woman, high fashion 2000s fashion.
% An astronaut riding a horse on the moon, oil painting by Van Gogh.
% Full body shot, a French woman, photography, French streets.
% A boy and a girl fall in love.
% An abstract portrait of a pensive face, rendered in cool shades of blues, purples, and grays.
% Cute boy, hair looking up to the stars, snow, beautiful lighting, painting style by Abe Toshiyuki.

%% visualization-c
%% prompt 13-21
% A table with a light on over it.
% A library filled with warm yellow light.
% A villa standing on a hill.
% A train crossing a bridge over a canyon.
% A bridge stretching over a calm river.
% A temple surrounded by flowers.

\end{document}